\def\withappendix{}
\def\arxivversion{}

\documentclass[10pt,twocolumn,letterpaper]{article}

\usepackage[pagenumbers]{cvpr}%

\usepackage{multirow}%

\definecolor{cvprblue}{rgb}{0.21,0.49,0.74}
\usepackage[pagebackref,breaklinks,colorlinks,allcolors=cvprblue]{hyperref}

\title{Trac3R: Pose-Grounded 3D Object Generative Reconstruction via Point Tracing}

\author{Minghan Zhu$^{1,2}$ \qquad Michael Posa$^{1}$\\
$^{1}$University of Pennsylvania \qquad $^{2}$University of Michigan\\
{\tt\small minghanz@umich.edu, posa@engineering.upenn.edu}
}

\begin{document}
\maketitle
\begin{abstract}
Estimating the complete geometry and pose of objects from an image is a fundamental capability for 3D scene understanding. Despite significant progress in generative shape reconstruction, the problem of grounding generation in observation for accurate poses remains challenging. We solve this joint shape-and-pose estimation problem by augmenting an image-to-3D object shape generation model to find correspondences between the complete shape and a partial observation, represented as a point cloud from a depth sensor or a depth-prediction neural network, so that the generated shape can be grounded in the observation frame. We amortize the correspondence-finding task into the shape generation process itself, by injecting the conditional point cloud as tracers to the shape latents, and observe where they end up in the fully generated shape. This leads to a simple architecture that is trained with a single shape generation objective, achieving state-of-the-art accuracy and efficiency. We demonstrate the wide applicability of our method on synthetic and real-world object-level and scene-level reconstruction tasks. 
\ifdefined\arxivversion
Project page: \url{https://trac3r-project.github.io}.
\fi
\end{abstract}
\begin{figure}[t]
\centering
\includegraphics[width=\columnwidth]{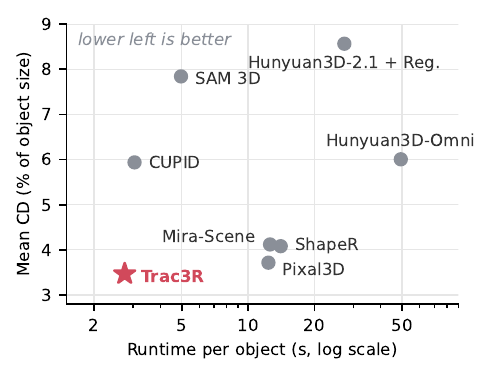}
\caption{Accuracy against runtime. Vertical axis: Chamfer distance (CD) between the predicted shape at the predicted pose and the ground truth, in \% of the longest side of the ground-truth object, averaged over the Toys4K~\cite{stojanov2021using}, GSO~\cite{downs2022googlescannedobjectshighquality} and OmniObject3D~\cite{wu2023omniobject3d} datasets with ground-truth and MoGe-2~\cite{wang2026moge} depth (\cref{tab:synthetic}). Horizontal axis: seconds per object at batch size 1 on one RTX PRO 6000 GPU, for shape and pose (appearance stages excluded if applicable); each method uses its released default settings. Trac3R is the only method that achieves both high accuracy and low runtime. See \cref{sec:exp_setup} for more details.}
\label{fig:teaser}
\end{figure}

\section{Introduction}
\label{sec:intro}

Most scenes can be approximated as a set of objects. Each object has a complete shape, and a camera sees one side of it placed at a certain pose in the scene. The complete shapes and poses of the objects together form a basic geometric state of the scene, and estimating them from a single image is a core problem in 3D vision. This state lets a physics simulator replicate a real scene, a robot reason about grasps and contacts on surfaces it cannot see, and an augmented-reality system place virtual content among real objects. 

Shape and pose have mostly been studied separately. Image-to-3D generative models~\cite{hunyuan3d2025hunyuan3d21,xiang_structured_2025}, trained on large collections of 3D assets, now produce high-quality complete shapes from one image of an unseen object. However, they generate each shape in its own canonical frame, with no pose relative to the observation. Object pose estimators~\cite{wen2024foundationpose} do the opposite: they align a shape to the observation, typically by correspondence matching, but they need the shape as an input, in the form of a CAD model, reference views, or a category-level prior. This assumption limits them in open-world settings, where the objects are not known in advance. A straightforward fix is to pass the generated shape to a pose estimator, but the two problems are then solved by separate models, and the pose stage does not benefit from the knowledge about the object that the generator has learned.

Recent work therefore estimates pose inside the shape generator, by regressing it~\cite{chen2026sam}, or by putting the shape in correspondence with the image pixels~\cite{huang2026cupid,li2026pixal3d,sun2026mira} or with an input point cloud~\cite{siddiqui2026shaper,hunyuan3d2025omni} (\cref{sec:related}). Regression needs realistic scenes with annotated object poses, which are hard to obtain at scale. In comparison, correspondence-based methods train on object-centric data that is already available at scale. However, two gaps remain. First, pixel-grounded methods use the depth only after generation, if at all. Second, point-grounded methods generate in the frame of the conditioning points, where the shape appears at an arbitrary orientation and offset and can extend beyond the bounded volume of the generator, which hurts the shape quality.

We follow the correspondence route and close both gaps: we establish the correspondences in 3D, between the complete shape in its canonical frame and a partial point cloud of the object that comes from a depth sensor or a monocular point-map predictor~\cite{wang2026moge}. Our key observation is that a conditional shape generator must already relate its condition to its output for the condition to be effective. We therefore do not add a module that searches for correspondences beside or after the generation. We make this relation explicit inside the generation process itself, by \emph{point tracing}.

\begin{figure*}[t]
\centering
\includegraphics [width=\textwidth]{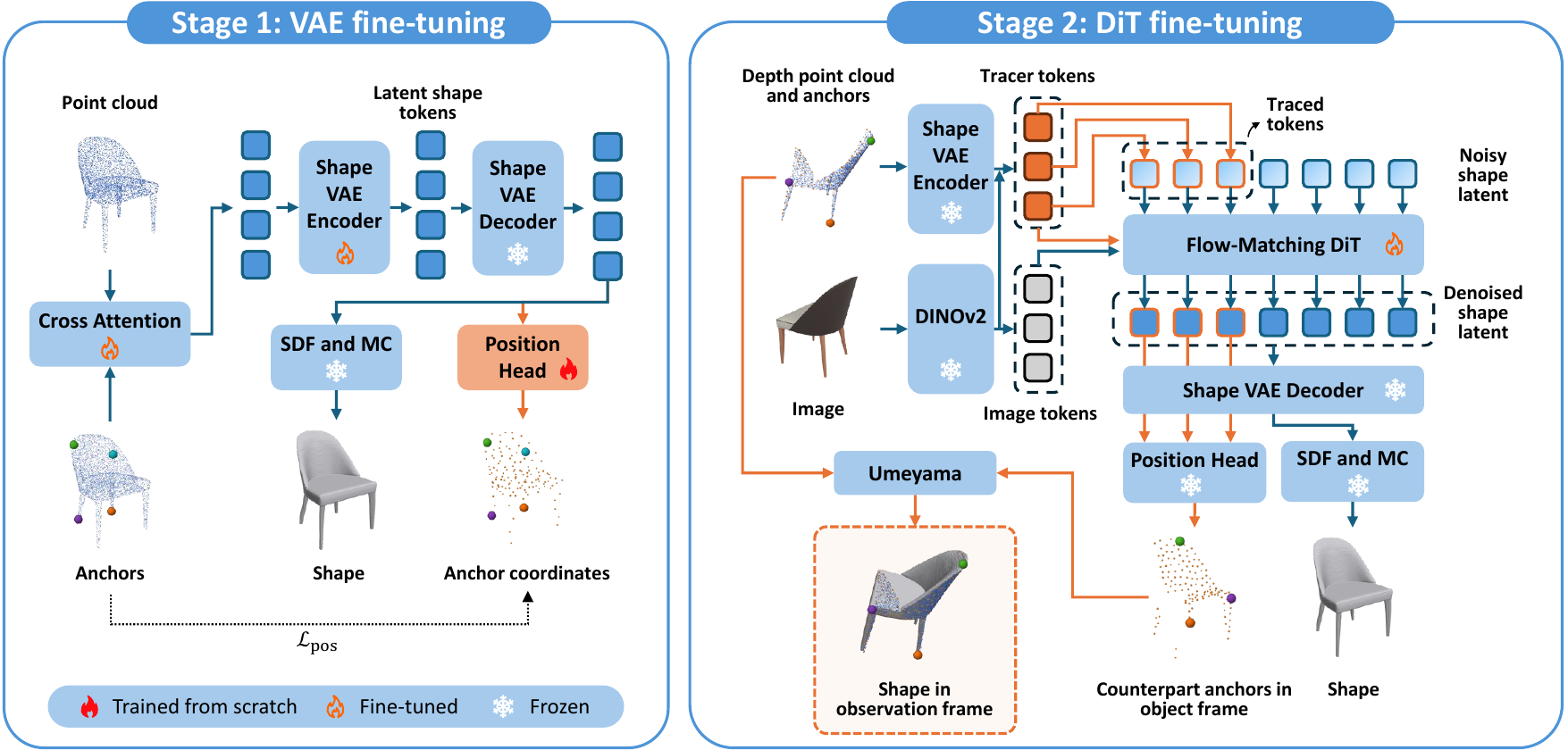}
\caption{Method overview. Orange arrows highlight the paths that Trac3R adds to the pretrained model to realize pose-grounded generation. \textbf{Stage~1:} a position head learns to predict the anchor (colored points) of each shape token. MC stands for marching cubes. \textbf{Stage~2:} tracers (orange) encoded from the observed anchors are added to the traced tokens and condition the DiT. The position head reads the counterparts of the observed anchors from the traced tokens, and an Umeyama fit to these pairs gives the pose. The numbers of tokens and anchors are for illustration only; the model uses 512 shape tokens, of which 128 are traced.}
\label{fig:method}
\end{figure*}

Our method, Trac3R, builds on a VecSet~\cite{zhang_3dshape2vecset_2023}-type generator~\cite{hunyuan3d2025hunyuan3d21}, which represents a shape as a set of latent tokens, each originating from a point on its surface. We bring the observed points into the same latent space as tokens, which we call \emph{tracers}, inject them into the shape latents, and let the generation carry them along. As the complete shape forms in the canonical object frame, each tracer settles where its observed point belongs on that shape, and we read out this position (\cref{fig:method}). Every tracer thus links a point of the observation to a point of the generated shape, and a closed-form fit to these links yields the pose. Because the correspondences arise from generating the shape itself, we obtain Trac3R by fine-tuning a pretrained generator on its original shape generation objective, with no matching module and no correspondence loss.

In our experiments, a single Trac3R model ranks among the top across synthetic and real-world benchmarks covering object and scene reconstruction tasks, with depth coming from ground truth, depth sensors, or a point map prediction model~\cite{wang2026moge}, while being the fastest of all compared methods, highlighting the generalization and efficiency of our approach. 

In summary, our contributions are as follows:
\begin{itemize}
    \item We propose point tracing, a simple mechanism that embeds point correspondence into a shape generation model: observed points are injected as tracers into the shape latents and read out on the generated shape.
    \item We build Trac3R, a joint object shape and pose estimation model with state-of-the-art accuracy and efficiency (\cref{fig:teaser}).
    \item We demonstrate wide applicability with a single model, from object reconstruction in real-world tabletop captures to indoor and outdoor scene reconstruction, which spans the perception domains of robot manipulation, augmented reality and real-to-sim transfer.
\end{itemize}

\section{Related Work}
\label{sec:related}

\paragraph{3D generation from a single image.}
Early methods distill 2D diffusion priors into a 3D representation~\cite{poole2022dreamfusion}, generate consistent novel views that are then fused~\cite{liu2023zero,long2024wonder3d}, or regress a 3D representation in one forward pass~\cite{hong2024lrm,tochilkin2024triposr,xu2024instantmesh}. Current methods train a diffusion model directly on a latent of 3D shapes, either an unordered set of tokens~\cite{zhang_3dshape2vecset_2023,zhang2024clay,li2025triposg,zhao2025hunyuan3d,hunyuan3d2025hunyuan3d21} or a structured grid~\cite{xiang_structured_2025}, and produce detailed shapes of unseen objects. Two properties of these models matter here. The shape is generated in the canonical frame of the training assets, so it has no pose in the observation. And the condition enters through cross-attention, so the relation between the input and the generated shape stays implicit~\cite{li2026pixal3d}. Trac3R makes this relation explicit for observed points, so that it gives the pose.

\paragraph{Object pose estimation.}
Pose estimators align an object model to the observation, either by direct regression~\cite{xiang2017posecnn,wang2019densefusion} or through dense correspondences, such as predicted object coordinates~\cite{park2019pix2pose,wang2021gdr,wang2019normalized}. The model is a known instance, a category prior~\cite{zhang2023genpose,zhang2024omni6dpose,zhang2026f2g}, a CAD model or reference views given at test time~\cite{labbe2022megapose,sun2022onepose,wen2024foundationpose,lin2024sam,nguyen2024gigapose}, or a reconstruction from an RGB-D video~\cite{wen2023bundlesdf,lin2026point2pose}. For an unseen object, the shape can come from a generator~\cite{lee2025any6d,agarwal2025scenecomplete}, but a second model, a pose estimator or a point-set registration~\cite{rusu2009fast,qin2023geotransformer}, must then match a generated shape to a partial and noisy observation without knowing how the shape was generated. In Trac3R, the model that produces the shape also produces the correspondences.

\paragraph{Scene reconstruction from a single image.}
Scenes are reconstructed object by object, from early layout and CAD-retrieval methods~\cite{nie2020total3dunderstanding,kuo2020mask2cad,gumeli2022roca} to pipelines that generate each object and place it with depth or a pose model~\cite{ardelean2025gen3dsr,zhou2024zero,yao2025cast,shi2026scenemaker}. Scene generators produce all objects together, so that their arrangement is part of the output~\cite{huang2025midi,meng2026scenegen,ling2026scene}, but they learn the arrangement from scene datasets, which are much smaller than object collections~\cite{sun2026mira}. Trac3R places each object from its own correspondences and is trained on objects only, so the same model applies to scenes without scene-level training.

\paragraph{Pose-grounded shape generation.}
Closest to our work is a recent line of generators that also place the shape in the observation. SAM~3D~\cite{chen2026sam} regresses the pose, a global quantity that depends on the camera and the scene layout rather than on the object alone. Pixel-relating methods ground the pose in correspondences instead: CUPID~\cite{huang2026cupid} and Mira-Scene~\cite{sun2026mira} generate a 3D--2D correspondence map beside the canonical shape, and Pixal3D~\cite{li2026pixal3d} generates the shape aligned to the pixels. However, they use the point map only after the generation, so the observed geometry guides neither the shape nor the correspondences. Point-conditioned methods~\cite{siddiqui2026shaper,hunyuan3d2025omni,xia2026points,hu2026axolotl3d} do use the points during generation, but generate the shape in the frame of the points. The shape then leaves the canonical frame and can extend beyond the bounded volume of the generator, which hurts its quality. Trac3R closes both gaps: the points condition the generation, and the shape stays in the canonical frame. It also needs no correspondence map beside the shape.

\paragraph{Correspondence from generative models.}
Generative models hold correspondences that they are not trained for. The features of image diffusion models match semantic parts between images~\cite{tang2023emergent,zhang2023tale}, and Rectified Point Flow~\cite{sun2026rectified} recovers part poses by generating the assembled positions of the given points. In these works, the correspondences are either matched from features after generation, or are between parts that are given. Trac3R relates a partial observation to a complete shape that is generated at the same time, and the correspondence is part of the generated latent.

\section{Method}
\label{sec:method}

\providecommand{\Enc}{\mathrm{Enc}}
\providecommand{\Dec}{\mathrm{Dec}}
\providecommand{\Pos}{\mathrm{Pos}}
\providecommand{\tr}{\mathrm{tr}}

The input is an RGB image $I$ of an object with its mask, and a partial point cloud $\mathcal{O}$ of the visible surface, back-projected from the masked depth. We call the frame of $\mathcal{O}$ the \emph{observation frame}. The output is the complete shape of the object in the canonical \emph{object frame} of the generator, and a similarity transform $T\in\mathrm{Sim}(3)$ that maps the object frame to the observation frame. Throughout, $\mathbf{s}$ is a point on the complete shape in the object frame, $\mathbf{o}$ is a point of the observation in the observation frame, $\mathbf{z}$ is a latent token, and a hat marks a prediction.

\Cref{fig:method} shows the two stages of Trac3R. Stage~1 (\cref{sec:method_head}) makes each shape token preserve the position on the shape that it comes from. Stage~2 (\cref{sec:method_tracing}) injects observed points into the generation as tracers. The positions that the traced tokens report after denoising give the pose in closed form (\cref{sec:method_pose}).

\subsection{Preliminaries: VecSet shape generation}
\label{sec:method_prelim}

We build on Hunyuan3D-2.1~\cite{hunyuan3d2025hunyuan3d21}, which follows the VecSet design~\cite{zhang_3dshape2vecset_2023}: a shape VAE and a flow-matching~\cite{lipman2023flow} diffusion transformer (DiT) on its latent tokens.

\paragraph{Shape VAE.}
Let $\mathcal{S}$ be a dense point cloud with normals, sampled from the surface of a shape. The encoder selects $N$ points $\mathbf{s}_1,\dots,\mathbf{s}_N\in\mathcal{S}$ by farthest point sampling (FPS). We call them \emph{anchors}. Each anchor attends to $\mathcal{S}$ by cross-attention, and self-attention layers follow. The result is one latent token per anchor,
\begin{equation}
\mathbf{z}_i=\Enc(\mathbf{s}_i\mid\mathcal{S}),\qquad i=1,\dots,N ,
\label{eq:encode}
\end{equation}
and the token set $Z=\{\mathbf{z}_i\}_{i=1}^{N}$ is the latent of the shape. The decoder processes $Z$ with self-attention layers to get a decoded token per anchor $\mathbf{h}_i=\Dec(\mathbf{z}_i\mid Z)$. A 3D location then attends to the decoded tokens $H=\{\mathbf{h}_i\}_{i=1}^{N}$ to give its signed distance, and marching cubes~\cite{lorensen1987marching} extracts the mesh.

\paragraph{Generator.}
The DiT $v_\theta$ generates $Z$ from noise, conditioned on DINOv2~\cite{oquab2024dinov2} tokens of the image $I$ through cross-attention. With $Z_t=tZ+(1-t)\boldsymbol{\epsilon}$, $\boldsymbol{\epsilon}\sim\mathcal{N}(\mathbf{0},\mathbf{I})$ and $t\in[0,1]$, the training objective is
\begin{equation}
\mathcal{L}_{\mathrm{FM}}=\mathbb{E}_{Z,\boldsymbol{\epsilon},t}\,\big\|\,v_\theta(Z_t,t\mid I)-(Z-\boldsymbol{\epsilon})\,\big\|^2 .
\label{eq:fm}
\end{equation}

\paragraph{Three properties that we use.}
Our method follows from three properties of this design.
\begin{itemize}[leftmargin=*,labelindent=0pt,align=left,font=\bfseries,widest=P2]
    \item[P1] \emph{A token originates from an anchor.} Each token $\mathbf{z}_i$ is created from one surface point, its anchor $\mathbf{s}_i$.
    \item[P2] \emph{The latent of a shape is not unique.} It depends on the choice of the anchors. Each FPS draw gives a different token set, and all of them decode to the same shape. The target of the generator is therefore a choice, not a given.
    \item[P3] \emph{The tokens have no order.} The DiT has no positional embedding and is permutation-equivariant, so the index $i$ of a token has no meaning.
\end{itemize}
P1 lets a token stand for a point of the shape (\cref{sec:method_head}). P2 and P3 let us design which points the generated tokens stand for, and which token stands for which point (\cref{sec:method_tracing}).

\subsection{Stage 1: VAE fine-tuning with a position head}
\label{sec:method_head}

By P1, each token has an anchor, but the pretrained VAE gives no way to read the anchor back from the token. We add a \emph{position head}: a three-layer MLP on each decoded token $\mathbf{h}_i$, which is trained with an L2 loss to predict the anchor coordinate $\mathbf{s}_i$ in the object frame. 
\begin{equation}
\mathcal{L}_{\mathrm{pos}}=\frac{1}{N}\sum_{i=1}^{N}\big\|\,\Pos(\mathbf{h}_i)-\mathbf{s}_i\,\big\|^2 .
\label{eq:pos}
\end{equation}

We fine-tune the encoder with LoRA~\cite{hu2021lora} together with the head, so that the position information is preserved in the latent tokens. The complete Stage~1 objective is
\begin{equation}
\mathcal{L}_{1}=\mathcal{L}_{\mathrm{pos}}+\lambda_{\mathrm{sdf}}\,\mathcal{L}_{\mathrm{sdf}}+\lambda_{\mathrm{KL}}\,\mathcal{L}_{\mathrm{KL}} ,
\label{eq:stage1}
\end{equation}
where $\mathcal{L}_{\mathrm{sdf}}$ and $\mathcal{L}_{\mathrm{KL}}$ are the reconstruction and KL terms of the original VAE. They preserve the shape compression capabilities and keep the latent distribution close to the pretrained DiT. After Stage~1, the VAE and the position head are frozen.

The head reads the position from a single token. It is therefore valid for any token in the shape latent space, including a token that the DiT generates.

\subsection{Stage 2: DiT fine-tuning for point tracing}
\label{sec:method_tracing}

We select $K$ anchors on the observation, $\mathbf{o}_1,\dots,\mathbf{o}_K\in\mathcal{O}$, by FPS. We call them the \emph{observed anchors}, and we want to know where they are on the generated shape. We make it part of what the generator produces: for each $\mathbf{o}_k$, the generated latent must contain an associated token $\mathbf{z}_k$ standing for the same point on the complete shape. The non-uniqueness of the shape latents for a given shape (P2) and the permutation equivariance of the latents (P3) allow us to build this correspondence into the DiT's training data, so that correspondence is learned via shape generation.

\paragraph{Tracers: the observed anchors as tokens.}
First, we bring the observed anchors into the shape latent space. We normalize $\mathcal{O}$ to the bounding-box size that the VAE expects, and encode the $K$ observed anchors with the frozen encoder of Stage~1:
\begin{equation}
\mathbf{z}^{\tr}_k=\Enc(\mathbf{o}_k\mid\mathcal{O}),\qquad k=1,\dots,K .
\label{eq:tracer}
\end{equation}
The normals that the encoder expects are estimated from the depth map: the normal at a pixel is the normalized cross product of the central differences of the back-projected points along the two image axes, oriented toward the camera. We call $\mathbf{z}^{\tr}_k$ a \emph{tracer}. It describes the surface around $\mathbf{o}_k$ as it is observed, in the observation frame.

\paragraph{Choosing the target latent (P2).}
Now, we design the target latent to train the DiT to represent the complete shape with correspondences to the observed anchors. Given an observed anchor $\mathbf{o}_k$ and the ground-truth pose $T^\ast$, we find its \emph{counterpart} $\mathbf{s}_k$ on the complete shape $\mathcal{S}$, which is the nearest neighbor to ${T^\ast}^{-1}(\mathbf{o}_k)$. We use the $K$ counterparts as the first $K$ anchors of the shape, and let FPS continue from them to select the other $N-K$. Such a set of anchors still encodes to a regular latent of the shape, but $K$ of the latent tokens $\mathbf{z}_k$ each correspond to a tracer $\mathbf{z}^{\tr}_k$ in the observation. 

\paragraph{Binding each tracer to a token (P3).}
The target latent now contains one counterpart token for each tracer, but the DiT cannot yet tell which token that is. By P3, the DiT treats all $N$ tokens alike: sampling starts from i.i.d.\ noise, and nothing distinguishes token $k$ from any other token. We therefore tag the tokens, by adding tracer $k$ to the noisy token $k$ at the input of the DiT, at every denoising step,
\begin{equation}
\mathbf{z}_{t,k}\;\leftarrow\;\mathbf{z}_{t,k}+\mathrm{Proj}\big(\mathbf{z}^{\tr}_k\big),\qquad k=1,\dots,K ,
\label{eq:inject}
\end{equation}
where $\mathbf{z}_{t,k}$ is token $k$ of the noisy latents $Z_t$ after the input embedding, and $\mathrm{Proj}$ is a linear layer. We call the first $K$ tokens the \emph{traced tokens}.

The tag breaks the permutation symmetry among the tokens: it makes the shape latent token $k$ identifiable with tracer $k$. In every training sample, the token that receives tracer $\mathbf{z}^{\tr}_k$ at the input has the counterpart token $\mathbf{z}_k$ as its target at the output. The other $N-K$ tokens carry no tag. They stay interchangeable, as in the pretrained generator. Since the $N$ tokens together are still a valid latent of the shape, the DiT can still generate the complete shape, but now it also doubles as a correspondence solver: the traced tokens are denoised to the tracers' counterparts in the complete shape.

The tag contains two auxiliary inputs. Each traced token receives the DINOv2 feature at the pixel of $\mathbf{o}_k$, and each token receives a learned embedding of its role, traced or not. In addition, the tracers join the image tokens as the context of the cross-attention, so that all $N$ tokens see the observation. All new layers are zero-initialized, so the fine-tuning starts at the pretrained generator.

\paragraph{Training.}
We fine-tune the DiT with the flow-matching loss of \cref{eq:fm}, with no pose loss or correspondence loss. We drop the image and the tracers at random during training for classifier-free guidance~\cite{ho2022classifier}.

\subsection{Pose from the traced tokens}
\label{sec:method_pose}

At test time, we generate a latent $\hat{Z}$ from the image and the tracers, and decode it, $\hat{\mathbf{h}}_i=\Dec(\hat{\mathbf{z}}_i\mid\hat{Z})$. The decoded tokens give the complete mesh in the object frame. On the traced tokens, the position head gives the predicted counterpart of each observed anchor, $\hat{\mathbf{s}}_k=\Pos(\hat{\mathbf{h}}_k)$ for $k\le K$. The pose is the least-squares fit to the $K$ pairs,
\begin{equation}
\hat{T}=\operatorname*{arg\,min}_{T\in\mathrm{Sim}(3)}\;\sum_{k=1}^{K}\big\|\,T(\hat{\mathbf{s}}_k)-\mathbf{o}_k\,\big\|^2 ,
\label{eq:pose}
\end{equation}
which has the closed-form solution of Umeyama~\cite{umeyama1991least}. Beyond the generation of the shape, the pose costs one MLP evaluation on $K$ tokens and one $3\times3$ singular value decomposition. The shape itself is generated in the object frame, as in the pretrained generator, so its quality does not depend on the pose of the object in the image.

\section{Experiments}
\label{sec:exp}

\subsection{Experimental setup}
\label{sec:exp_setup}

\paragraph{Task and benchmarks.}
A test case is one object in one image. The input is the RGB image, the mask of the visible part of the object, the camera intrinsics, and a depth map. The output is a complete mesh of the object, placed in the camera frame. The depth is one of three kinds: the rendered ground-truth depth (GT); the depth of a sensor; or the point map that MoGe-2~\cite{wang2026moge} predicts from the RGB image. The global scale of a predicted point map is uncertain, so we fit it to a reference depth by scaling and shifting. Sensor and predicted depth are screened for outliers.

We train a single model and evaluate it on four benchmarks. On synthetic objects (\cref{sec:exp_synthetic}), rendered alone with no occlusion, we compare GT depth with predicted depth. The HANDAL dataset~\cite{guo2023handal} (\cref{sec:exp_handal}) has real household objects with little or no occlusion, and we use predicted depth. The BOP benchmark~\cite{hodan2018bop,hodan2024bop} (\cref{sec:exp_bop}) has sensor depth and objects in clutter and piles, with substantial mutual occlusion. The BlendSwap dataset~\cite{blendswap} (\cref{sec:exp_scene}) has synthetic indoor and outdoor scenes with GT depth, where occlusion arises naturally from the layout of each scene. 
Together, these benchmarks cover the settings of several downstream applications: object perception on devices without a depth sensor for augmented reality; robot manipulation and bin picking with a depth sensor; and scene reconstruction for real-to-sim transfer. Within a benchmark, all methods receive the same image, mask, intrinsics and depth.

\paragraph{Baselines.}
We compare with the three groups of methods of \cref{sec:intro}: (i)~\emph{direct regression}, SAM~3D~\cite{chen2026sam}; (ii)~\emph{pixel correspondence}, CUPID~\cite{huang2026cupid}, Pixal3D~\cite{li2026pixal3d} and Mira-Scene~\cite{sun2026mira}; and (iii)~\emph{point-conditioned generation}, ShapeR~\cite{siddiqui2026shaper} and Hunyuan3D-Omni~\cite{hunyuan3d2025omni}. SAM~3D takes the full-frame image and point map, together with the object mask. The methods of groups (ii) and (iii), like ours, take an object-centric crop. 

We add a two-stage baseline, (iv)~\emph{generate, then register} (Hunyuan3D-2.1 + Reg.): Hunyuan3D-2.1~\cite{hunyuan3d2025hunyuan3d21}, the generator family that Trac3R is fine-tuned from, generates the shape in its canonical frame, and we register it to the observed point cloud by FPFH~\cite{rusu2009fast} matching with RANSAC, followed by multi-start Sim(3) ICP.

All baselines run with their released checkpoints and default settings. On the real benchmarks we add two references that are given the ground-truth shape, and are therefore not ranked: \emph{GT mesh + Reg.}\ registers the ground-truth mesh to the depth with the same registration as Hunyuan3D-2.1 + Reg., and FoundationPose~\cite{wen2024foundationpose} estimates the pose of the ground-truth mesh. They show how much of the error comes from placing a perfect shape.

\begin{figure*}[t]
\centering
\includegraphics[width=\textwidth]{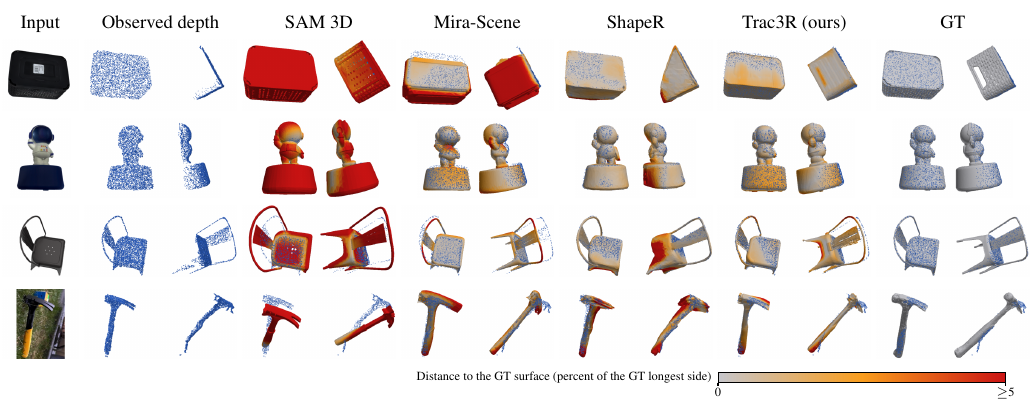}
\caption{Rows 1--3: synthetic objects with GT depth (from GSO, OmniObject3D and Toys4K). Row 4: a real object with MoGe-2 depth (from HANDAL). Left: the input image; for HANDAL we keep the real background for visualization, but our method reads the masked input. Second column: the observed point cloud. Other columns: every mesh placed with its predicted pose, with the observed point cloud in blue. Each mesh is colored by the distance of its vertices to the GT surface, in percent of the longest side of the GT object (gray: 0, red: 5 or more). Each column shows the camera view and a view turned by 70$^\circ$ around the object.}
\label{fig:qual_synthetic}
\end{figure*}

\begin{table}[t]
\centering
\footnotesize
\setlength{\tabcolsep}{4pt}
\begin{tabular}{l ccc}
\toprule
Method & \multicolumn{1}{c}{Toys4K~\cite{stojanov2021using}} & \multicolumn{1}{c}{GSO~\cite{downs2022googlescannedobjectshighquality}} & \multicolumn{1}{c}{OmniObject3D~\cite{wu2023omniobject3d}} \\
\midrule
\multicolumn{4}{l}{\textit{GT depth}} \\
SAM 3D~\cite{chen2026sam} & 7.16/7.75 & 7.51/8.13 & 7.15/7.56 \\
CUPID~\cite{huang2026cupid} & 3.66/5.49 & 4.33/5.97 & 4.38/5.95 \\
Pixal3D~\cite{li2026pixal3d} & \underline{2.41}/\underline{2.85} & 3.21/4.34 & 3.16/3.97 \\
Mira-Scene~\cite{sun2026mira} & 2.81/3.49 & 3.41/4.24 & 3.65/4.46 \\
ShapeR~\cite{siddiqui2026shaper} & 2.42/2.86 & \textbf{2.70}/\textbf{3.00} & \underline{3.12}/\underline{3.76} \\
HY3D-Omni~\cite{hunyuan3d2025omni} & 3.83/4.48 & 3.87/4.56 & 4.91/5.44 \\
HY3D-2.1~\cite{hunyuan3d2025hunyuan3d21} + Reg. & 3.04/6.24 & 4.61/8.47 & 4.44/8.04 \\
\cmidrule(lr){1-4}
Trac3R (ours) & \textbf{2.26}/\textbf{2.79} & \underline{2.77}/\underline{3.44} & \textbf{3.02}/\textbf{3.68} \\
\midrule
\multicolumn{4}{l}{\textit{MoGe-2 depth}} \\
SAM 3D~\cite{chen2026sam} & 7.16/7.83 & 7.69/8.19 & 7.16/7.58 \\
CUPID~\cite{huang2026cupid} & 4.05/5.72 & 4.70/6.23 & 4.94/6.26 \\
Pixal3D~\cite{li2026pixal3d} & \textbf{2.44}/\textbf{2.89} & \underline{3.21}/\underline{4.32} & \textbf{3.16}/\underline{3.95} \\
Mira-Scene~\cite{sun2026mira} & 2.97/3.68 & 3.56/\underline{4.32} & \underline{3.82}/4.53 \\
ShapeR~\cite{siddiqui2026shaper} & 3.53/4.38 & 3.71/4.60 & 4.58/5.89 \\
HY3D-Omni~\cite{hunyuan3d2025omni} & 5.61/6.46 & 5.64/6.90 & 7.01/8.19 \\
HY3D-2.1~\cite{hunyuan3d2025hunyuan3d21} + Reg. & 5.06/8.58 & 6.75/10.29 & 6.01/9.76 \\
\cmidrule(lr){1-4}
Trac3R (ours) & \underline{2.62}/\underline{3.28} & \textbf{3.17}/\textbf{3.79} & \textbf{3.16}/\textbf{3.83} \\
\bottomrule
\end{tabular}
\caption{Synthetic objects with GT and MoGe-2 depth, one block of rows per depth input. CD in \% of object size, median\,/\,mean. \textbf{Bold:} best; \underline{underline}: second best, per column within each depth block. HY3D: Hunyuan3D.}
\label{tab:synthetic}
\end{table}

\paragraph{Metric.}
The main metric on all four benchmarks is the Chamfer distance (CD) between the predicted mesh at the predicted pose and the ground-truth mesh at the ground-truth pose,
$\mathrm{CD}(A,B)=\tfrac12\big(\tfrac{1}{|A|}\sum_{a\in A}\min_{b\in B}\|a-b\|+\tfrac{1}{|B|}\sum_{b\in B}\min_{a\in A}\|a-b\|\big)$,
with $A$ and $B$ sampled from the two surfaces.
It is low only if the shape is right and it is in the right place. We prefer it to rotation and translation errors, which are not well defined for symmetric objects and ignore the shape. On synthetic data we report the CD in \% of the longest side of the ground-truth object; on real data, we report CD in millimeters. We report the median and the mean over cases; lower is better. Metrics that are specific to one benchmark are defined in its section.

\begin{table}[t]
\centering
\footnotesize
\setlength{\tabcolsep}{2.5pt}
\resizebox{\columnwidth}{!}{%
\begin{tabular}{l c ccc}
\toprule
 & & \multicolumn{3}{c}{Box AUC $\uparrow$} \\
\cmidrule(lr){3-5}
Method & CD (mm) $\downarrow$ & @25 & @50 & @75 \\
\midrule
F2G-Pose~\cite{zhang2026f2g} & 11.23\,/\,13.87 & 17.4 & 2.8 & 0.0 \\
SAM 3D~\cite{chen2026sam} & 13.72\,/\,19.45 & 16.2 & 3.3 & 0.0 \\
CUPID~\cite{huang2026cupid} & 16.19\,/\,20.35 & 15.7 & 4.4 & 0.2 \\
Pixal3D~\cite{li2026pixal3d} & 8.66\,/\,12.85 & 25.6 & 9.0 & \underline{0.7} \\
Mira-Scene~\cite{sun2026mira} & 9.93\,/\,16.06 & 22.6 & 6.8 & 0.3 \\
ShapeR~\cite{siddiqui2026shaper} & \underline{7.57}\,/\,\underline{10.42} & \underline{30.5} & \textbf{12.3} & \textbf{0.8} \\
HY3D-2.1~\cite{hunyuan3d2025hunyuan3d21} + Reg. & 12.51\,/\,22.31 & 16.9 & 4.2 & 0.0 \\
HY3D-2.1~\cite{hunyuan3d2025hunyuan3d21} + FP~\cite{wen2024foundationpose} & 14.27\,/\,34.38 & 18.6 & 6.4 & 0.4 \\
\midrule
Trac3R (ours) & \textbf{7.10}\,/\,\textbf{9.70} & \textbf{31.9} & \underline{11.5} & \underline{0.7} \\
\midrule
{\color{gray}GT mesh + Reg.} & {\color{gray}10.20\,/\,15.48} & {\color{gray}22.4} & {\color{gray}7.5} & {\color{gray}0.8} \\
{\color{gray}GT mesh + FP~\cite{wen2024foundationpose}} & {\color{gray}3.67\,/\,6.46} & {\color{gray}56.5} & {\color{gray}41.1} & {\color{gray}17.2} \\
\bottomrule
\end{tabular}}
\caption{Real objects with MoGe-2 depth on HANDAL. CD in mm, median\,/\,mean. Box AUC@25/50/75 is normalized to 0--100. $\downarrow$: lower is better; $\uparrow$: higher is better. FP: FoundationPose.}
\label{tab:handal}
\end{table}

\paragraph{Implementation details.}
Both training stages use HY3D-Bench~\cite{hunyuan3d2026hy3d} objects that do not overlap with any test set. In Stage 1, we fine-tune the VAE on 10k objects, with LoRA~\cite{hu2021lora} on the encoder; this takes less than 12 hours on one B200 GPU. In Stage 2, we fine-tune the DiT on renders of 40k objects. The depth of each render is the MoGe-2 point map, fitted to the ground-truth shape. The observed point cloud has 4{,}096 points, of which 128 are sampled to be tracers. This stage takes less than 24 hours on two B200 GPUs. At test time we crop the image around the object mask and sample with 50 steps in denoising.

\begin{figure*}[t]
\centering
\includegraphics[width=\textwidth]{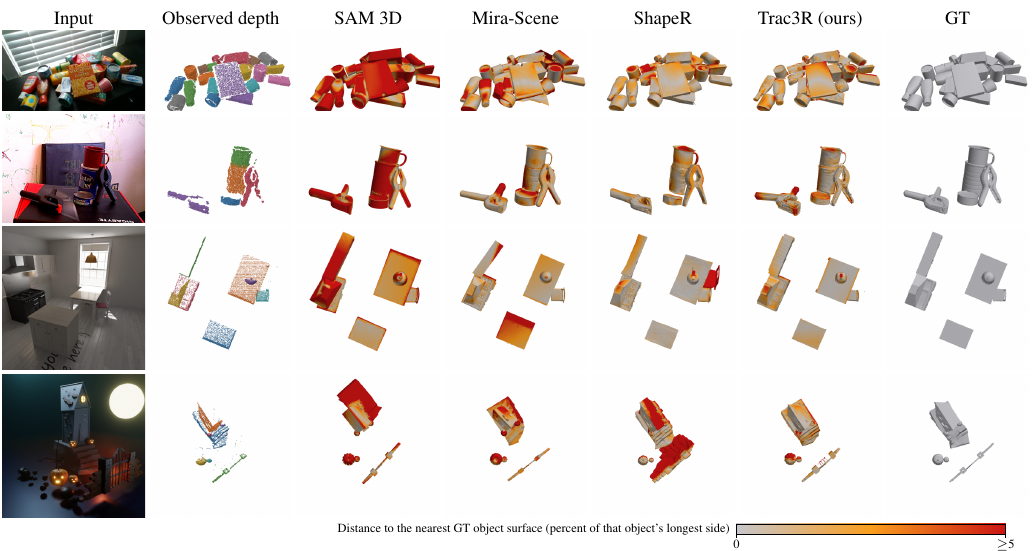}
\caption{Rows 1--2: real object piles with sensor depth (from HOPE and YCB-V). Rows 3--4: scenes with GT depth (from BlendSwap). Left: the input image. Second column: the observed point cloud of each object. Other columns: every placed mesh, colored by the distance of its vertices to the nearest GT object surface, in percent of the longest side of that GT object (gray: 0, red: 5 or more). BOP rows are seen from the camera; BlendSwap rows are seen from straight above to better visualize the scene layout. }
\label{fig:qual_frames}
\end{figure*}

\begin{table*}[t]
\centering
\footnotesize
\setlength{\tabcolsep}{3pt}
\resizebox{\textwidth}{!}{%
\begin{tabular}{l cccccccc}
\toprule
Dataset & YCB-V & LM-O & T-LESS & HB & HOPE & TUD-L & IC-BIN & ITODD \\
\textit{\scriptsize objects\,/\,cases} & {\scriptsize 21\,/\,496} & {\scriptsize 8\,/\,498} & {\scriptsize 30\,/\,497} & {\scriptsize 33\,/\,497} & {\scriptsize 28\,/\,500} & {\scriptsize 3\,/\,500} & {\scriptsize 2\,/\,499} & {\scriptsize 28\,/\,121} \\
\midrule
SAM 3D~\cite{chen2026sam} & 7.59\,/\,11.36 & 13.03\,/\,21.43 & 5.55\,/\,12.56 & 13.19\,/\,23.74 & 6.53\,/\,9.10 & 19.76\,/\,21.76 & 14.03\,/\,15.34 & 4.06\,/\,\underline{4.76} \\
CUPID~\cite{huang2026cupid} & 15.27\,/\,18.27 & 22.53\,/\,28.09 & 9.67\,/\,18.10 & 17.40\,/\,21.90 & 10.73\,/\,12.60 & 25.14\,/\,38.09 & 28.26\,/\,33.72 & 7.03\,/\,11.72 \\
Pixal3D~\cite{li2026pixal3d} & 5.45\,/\,7.40 & 11.36\,/\,12.82 & 4.53\,/\,9.33 & 7.93\,/\,10.37 & 5.77\,/\,7.49 & 14.88\,/\,17.33 & 13.02\,/\,15.09 & \underline{3.43}\,/\,6.30 \\
Mira-Scene~\cite{sun2026mira} & \underline{5.38}\,/\,\underline{6.44} & 9.41\,/\,11.66 & \underline{4.36}\,/\,8.81 & \underline{7.43}\,/\,9.58 & \underline{5.41}\,/\,6.71 & 15.05\,/\,16.93 & 12.28\,/\,15.53 & 3.61\,/\,4.91 \\
ShapeR~\cite{siddiqui2026shaper} & 6.82\,/\,7.83 & \underline{7.90}\,/\,\underline{9.81} & 4.64\,/\,\textbf{6.28} & 7.55\,/\,\underline{9.41} & 5.50\,/\,\underline{6.07} & \textbf{10.71}\,/\,\textbf{11.61} & \underline{11.35}\,/\,\underline{14.35} & \textbf{3.08}\,/\,\textbf{3.98} \\
\midrule
Trac3R (ours) & \textbf{4.68}\,/\,\textbf{5.85} & \textbf{7.36}\,/\,\textbf{9.00} & \textbf{3.84}\,/\,\underline{7.08} & \textbf{6.69}\,/\,\textbf{8.34} & \textbf{4.62}\,/\,\textbf{5.97} & \underline{10.78}\,/\,\underline{11.76} & \textbf{10.76}\,/\,\textbf{13.01} & 3.78\,/\,4.92 \\
\midrule
{\color{gray}GT mesh + Reg.} & {\color{gray}6.21\,/\,9.76} & {\color{gray}10.70\,/\,14.21} & {\color{gray}6.31\,/\,11.42} & {\color{gray}9.05\,/\,14.16} & {\color{gray}5.46\,/\,10.22} & {\color{gray}9.48\,/\,15.59} & {\color{gray}7.45\,/\,14.02} & {\color{gray}3.01\,/\,4.93} \\
{\color{gray}GT mesh + FP~\cite{wen2024foundationpose}} & {\color{gray}2.40\,/\,2.48} & {\color{gray}3.83\,/\,4.96} & {\color{gray}1.80\,/\,5.57} & {\color{gray}2.63\,/\,3.01} & {\color{gray}1.70\,/\,1.96} & {\color{gray}3.70\,/\,3.79} & {\color{gray}3.66\,/\,4.67} & {\color{gray}1.12\,/\,1.43} \\
\bottomrule
\end{tabular}}
\caption{Real objects with sensor depth on BOP. Second header row: number of objects and cases. CD in mm, median\,/\,mean.}
\label{tab:bop}
\end{table*}

\subsection{Synthetic objects with ground-truth and predicted depth}
\label{sec:exp_synthetic}

We start with isolated, unoccluded objects. From each of Toys4K~\cite{stojanov2021using}, GSO~\cite{downs2022googlescannedobjectshighquality} and OmniObject3D~\cite{wu2023omniobject3d}, we sample about 500 objects and render one view of each. We run every method twice: with the GT depth, and with the MoGe-2 point map fitted to the GT depth.

\Cref{tab:synthetic} shows the results. Trac3R is first in most columns and second in the rest. Compared with the other point-conditioned methods, Trac3R is more accurate and more robust to depth errors, with smaller error increase between ground truth depth and MoGe-2 depth. Pixel-matching methods are also robust to depth noise, but their overall accuracy is lower. The two-stage pipeline Hunyuan3D-2.1 + Reg.\ is less accurate than Trac3R, specifically with a larger median-mean gap, showing that the separate shape and pose estimation pipeline is more prone to catastrophic failures.
\Cref{fig:qual_synthetic} shows qualitative comparison among the methods. Trac3R produces both high-quality shapes and accurate placements.

Trac3R is also the fastest method (\cref{fig:teaser}): 2.8\,s per object on one RTX PRO 6000 GPU, against 3.1\,s (CUPID) to 49.4\,s (Hunyuan3D-Omni) for the baselines. The pose adds 1\,ms to the generation, because the correspondences are read from the generated tokens and the fit is closed-form.

\begin{table}[t]
\centering
\footnotesize
\setlength{\tabcolsep}{3pt}
\resizebox{\columnwidth}{!}{%
\begin{tabular}{l cccc}
\toprule
Method & CD $\downarrow$ & IoU-B $\uparrow$ & CD-S $\downarrow$ & F-Score-S $\uparrow$ \\
\midrule
MIDI~\cite{huang2025midi} & 28.57\,/\,102.31 & 0.111\,/\,0.200 & 0.1144\,/\,0.1227 & 0.545\,/\,0.592 \\
SAM 3D~\cite{chen2026sam} & 4.22\,/\,5.02 & 0.636\,/\,0.614 & 0.0021\,/\,0.0047 & \textbf{0.990}\,/\,\underline{0.967} \\
Pixal3D~\cite{li2026pixal3d} & 4.39\,/\,6.59 & 0.630\,/\,0.594 & 0.0050\,/\,0.0083 & 0.949\,/\,0.952 \\
Mira-Scene~\cite{sun2026mira} & \underline{3.27}\,/\,\textbf{3.74} & \textbf{0.771}\,/\,\textbf{0.741} & \underline{0.0014}\,/\,\textbf{0.0018} & 0.987\,/\,\textbf{0.982} \\
ShapeR~\cite{siddiqui2026shaper} & 5.16\,/\,10.27 & 0.604\,/\,0.531 & 0.0071\,/\,0.0109 & 0.926\,/\,0.904 \\
\midrule
Trac3R (ours) & \textbf{3.18}\,/\,\underline{3.91} & \underline{0.729}\,/\,\underline{0.709} & \textbf{0.0013}\,/\,\underline{0.0034} & \underline{0.989}\,/\,0.965 \\
\bottomrule
\end{tabular}}
\caption{Scenes with GT depth on BlendSwap (15 scenes, 88 objects), scored as placed, with no alignment. CD (\% of the object's longest side) and IoU-B per object; CD-S and F-Score-S per scene. Median\,/\,mean. }
\label{tab:scene}
\end{table}

\subsection{Real objects with predicted depth}
\label{sec:exp_handal}

HANDAL~\cite{guo2023handal} is a dataset of real captures of manipulable household objects. We use 914 cases: 92 objects of 7 categories, 10 frames each. HANDAL has no sensor depth; its ground-truth geometry comes from a NeRF reconstruction of each scene. Every method receives the MoGe-2 point map of the RGB image, fitted by scaling and shifting to the depth rendered from that reconstruction. We add F2G-Pose~\cite{zhang2026f2g}, a state-of-the-art category-level method that predicts pose, size and shape from RGB-D and evaluates on HANDAL, and a second two-stage pipeline, in which FoundationPose places the shape generated by Hunyuan3D-2.1 (Hunyuan3D-2.1 + FoundationPose). Besides the CD, we report the metric of the category-level pose literature, the box AUC of Omni6DPose~\cite{zhang2024omni6dpose}. For an IoU threshold $\tau$, the success rate is the fraction of cases whose oriented 3D box has an IoU above $\tau$ with the ground-truth box. AUC@$t$ is the area under the success rate as a function of $\tau$ from $t$ to 1, averaged over the categories; we report $t=0.25$, 0.50 and 0.75.

\Cref{tab:handal} shows the results. Trac3R has the lowest CD. On the box AUC, Trac3R and ShapeR are close. The other methods are clearly behind. Both two-stage pipelines have a heavy tail. Of the two references, GT mesh + Reg.\ is less accurate than Trac3R despite perfect shapes, showing the limitation of traditional feature matching. FoundationPose with the ground-truth mesh shows the gain that a known object model still gives.

\subsection{Real objects with sensor depth}
\label{sec:exp_bop}

In real-world applications, the sensor depth is metric but can be noisy and incomplete, and cluttered scenes add occlusion. We test this setting on eight datasets of the BOP benchmark~\cite{hodan2018bop,hodan2024bop}: YCB-V, LM-O, T-LESS, HB, HOPE, TUD-L, IC-BIN and ITODD, which cover household and industrial objects on tabletops and in bins. From each dataset we draw 500 cases (ITODD only has 121, so we use all), stratified over objects. The datasets are not equally diverse, as TUD-L and IC-BIN only have 3 and 2 objects, respectively (\cref{tab:bop}, second header row).

As shown in \cref{tab:bop}, Trac3R has the lowest median CD on six of the eight datasets and the lowest mean on five. achieving the best performance overall. It shows the robustness of Trac3R in realistic challenging scenarios with imperfect sensor depth and substantial occlusion. \Cref{fig:qual_frames} (rows 1--2) shows two object piles from HOPE and YCB-V wit the reconstruction from several methods, where Trac3R produces the most accurate shapes and placements. 

\subsection{Indoor and outdoor scenes}
\label{sec:exp_scene}

The last experiment addresses the task of reconstructing indoor and outdoor scenes. We use BlendSwap~\cite{blendswap} in the version released with Mira-Scene~\cite{sun2026mira}: 15 synthetic scenes with 88 objects, with GT depth. We add the scene generator MIDI~\cite{huang2025midi} as another baseline, which reads the full-scene image as input.
Besides the per-object CD, we report three metrics of MIDI: IoU-B, the volume IoU of the axis-aligned boxes of each predicted and ground-truth object (higher is better); and, on the merged scene scaled to a longest side of 1.9, CD-S, a squared Chamfer distance, and F-Score-S, the F-score at a distance threshold of 0.1. 

\Cref{tab:scene} shows the results, where Trac3R performs comparably with Mira-Scene and both leads ahead of the other methods. Mira-Scene is designed for scenes, whereas Trac3R is the same single-object model as in the other three experiments. It shows the generalization of Trac3R across different scenarios. Rows 3--4 of \cref{fig:qual_frames} show the reconstruction of one indoor and one outdoor scene in top-down view, demonstrating the accurate layout recovery of Trac3R.

We provide an ablation study and a more detailed analysis of the results in the appendix.

\section{Conclusion}
\label{sec:conclusion}

We presented Trac3R, which grounds a generated shape in its observation by point tracing. The method rests on one observation: a conditional shape generator already relates its condition to its output, so correspondences need not be searched for by a separate module. Injecting observed points as tracers and reading them out on the generated shape makes this relation explicit, at the cost of a light fine-tuning and a negligible runtime overhead. A single Trac3R model is accurate across object and scene benchmarks with ground-truth, sensor and predicted depth, and it is the fastest of the compared methods. This makes pose-grounded generation a practical perception front end for robot manipulation, augmented reality and real-to-sim transfer. Trac3R also has limitations. It relies on the given mask and depth and does not correct their errors. It generates each object independently, without reasoning about the relations between objects in a scene. Extending point tracing to multiple views and to joint generation of several objects is a promising direction.

\ifdefined\arxivversion
\section*{Acknowledgements}
This work was supported by an NSF CAREER Award under Grant No.~FRR-2238480, and the RAI Institute.
\fi

{
    \small
    \bibliographystyle{ieeenat_fullname}
    \bibliography{main}
}

\ifdefined\withappendix
\clearpage
\setcounter{figure}{0}
\setcounter{table}{0}
\renewcommand{\thetable}{S\arabic{table}}
\renewcommand{\thefigure}{S\arabic{figure}}
\renewcommand{\theHtable}{S\arabic{table}}
\renewcommand{\theHfigure}{S\arabic{figure}}

\makeatletter
\twocolumn[{%
  \centering
  \ifdefined\withappendix
    {\Large\textbf{Appendix}\\}
  \else
    {\Large\textbf{\thetitle}\\
    \vspace{0.5em}Appendix\\}
  \fi
  \vspace{1.0em}
  \begingroup
  \renewenvironment{figure*}[1][]{\def\@captype{figure}\par}{\par}%
  \begin{figure*}[t]
\centering
\includegraphics[width=\textwidth]{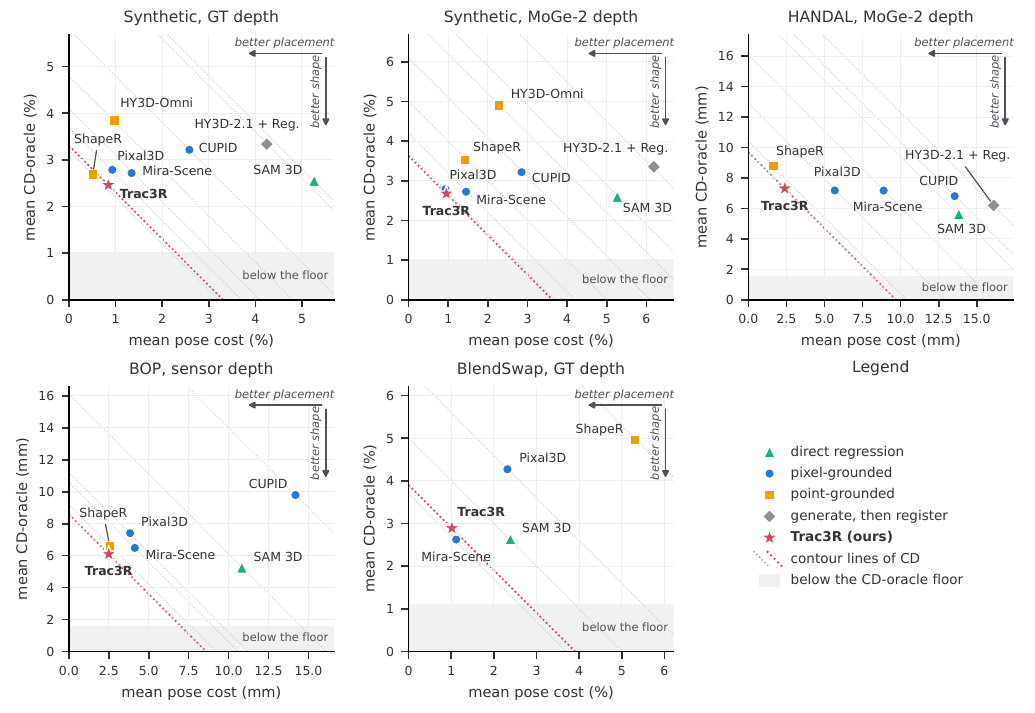}
\caption{\textbf{Shape error against placement error.} Each point is one method on one benchmark. The CD-oracle scores the shape alone. The pose cost measures the error caused by suboptimal pose estimation (see \cref{sec:appendix_tradeoff}). The two add up to the CD, so the dotted lines are contour lines of CD. The red line passes through Trac3R. The gray band marks the CD floor of a perfect shape. \Cref{tab:tradeoff} reports the numbers and the medians.}
\label{fig:tradeoff}
\end{figure*}
  \endgroup
  \vspace{1.5em}
}]
\makeatother
\appendix
\noindent This appendix is organized as follows:
\begin{itemize}
\item \Cref{sec:appendix_tradeoff} splits the error of each method into shape and pose, explains where the advantage of Trac3R comes from, and shows more qualitative results.
\item \Cref{sec:appendix_runtime} breaks down the runtime into stages.
\item \Cref{sec:exp_ablation} ablates the main design choices and the number of tracers.
\item \Cref{sec:appendix_fail} analyzes failure cases.
\item \Cref{sec:appendix_impl} gives further implementation details.
\end{itemize}

\section{Shape and pose error analysis}
\label{sec:appendix_tradeoff}

\begin{table*}[t]
\centering
\footnotesize
\setlength{\tabcolsep}{3pt}
\begin{tabular}{l l ccccc}
\toprule
 & & Synthetic & Synthetic & HANDAL & BOP & BlendSwap \\
Group & Method & {\scriptsize GT depth, \%} & {\scriptsize MoGe-2 depth, \%} & {\scriptsize MoGe-2 depth, mm} & {\scriptsize sensor depth, mm} & {\scriptsize GT depth, \%} \\
\midrule
\multicolumn{7}{c}{\textbf{CD-oracle (shape)}} \\
\midrule
\multirow{1}{*}{Generate, then register} & HY3D-2.1~\cite{hunyuan3d2025hunyuan3d21} + Reg. & 2.39\,/\,3.34 & 2.40\,/\,3.35 & \underline{4.91}\,/\,\underline{6.20} & -- & -- \\
\cmidrule(lr){1-7}
\multirow{1}{*}{Direct regression} & SAM 3D~\cite{chen2026sam} & \textbf{2.09}\,/\,\underline{2.55} & \textbf{2.11}\,/\,\textbf{2.60} & \textbf{4.68}\,/\,\textbf{5.62} & \textbf{3.92}\,/\,\textbf{5.25} & \underline{2.27}\,/\,\underline{2.64} \\
\cmidrule(lr){1-7}
\multirow{3}{*}{Pixel-grounded} & CUPID~\cite{huang2026cupid} & 2.38\,/\,3.22 & 2.38\,/\,3.22 & 5.43\,/\,6.80 & 7.91\,/\,9.79 & -- \\
 & Pixal3D~\cite{li2026pixal3d} & 2.36\,/\,2.79 & \underline{2.36}\,/\,2.79 & 5.40\,/\,7.18 & 6.03\,/\,7.41 & 3.25\,/\,4.27 \\
 & Mira-Scene~\cite{sun2026mira} & 2.35\,/\,2.72 & 2.37\,/\,2.73 & 5.50\,/\,7.17 & 5.18\,/\,6.48 & \textbf{2.18}\,/\,\textbf{2.63} \\
\cmidrule(lr){1-7}
\multirow{3}{*}{Point-grounded} & ShapeR~\cite{siddiqui2026shaper} & 2.42\,/\,2.69 & 3.17\,/\,3.53 & 6.51\,/\,8.78 & 5.77\,/\,6.59 & 4.16\,/\,4.96 \\
 & HY3D-Omni~\cite{hunyuan3d2025omni} & 3.41\,/\,3.85 & 4.15\,/\,4.91 & -- & -- & -- \\
 & Trac3R (ours) & \underline{2.14}\,/\,\textbf{2.46} & 2.37\,/\,\underline{2.68} & 5.64\,/\,7.30 & \underline{5.05}\,/\,\underline{6.10} & 2.53\,/\,2.89 \\
\cmidrule(lr){1-7}
 & {\color{gray}Floor} & {\color{gray}1.02} & {\color{gray}1.02} & {\color{gray}1.54} & {\color{gray}1.62} & {\color{gray}1.12} \\
\midrule
\multicolumn{7}{c}{\textbf{Pose cost (placement)}} \\
\midrule
\multirow{1}{*}{Generate, then register} & HY3D-2.1~\cite{hunyuan3d2025hunyuan3d21} + Reg. & 1.05\,/\,4.24 & 2.74\,/\,6.19 & 7.20\,/\,16.11 & -- & -- \\
\cmidrule(lr){1-7}
\multirow{1}{*}{Direct regression} & SAM 3D~\cite{chen2026sam} & 4.82\,/\,5.26 & 4.86\,/\,5.27 & 8.28\,/\,13.82 & 5.38\,/\,10.83 & 1.64\,/\,2.39 \\
\cmidrule(lr){1-7}
\multirow{3}{*}{Pixel-grounded} & CUPID~\cite{huang2026cupid} & 1.47\,/\,2.59 & 1.83\,/\,2.85 & 8.67\,/\,13.55 & 9.07\,/\,14.19 & -- \\
 & Pixal3D~\cite{li2026pixal3d} & \underline{0.26}\,/\,0.93 & \textbf{0.29}\,/\,\textbf{0.93} & 2.16\,/\,5.67 & 1.68\,/\,3.83 & 0.97\,/\,2.32 \\
 & Mira-Scene~\cite{sun2026mira} & 0.69\,/\,1.35 & 0.86\,/\,1.45 & 3.79\,/\,8.88 & 1.86\,/\,4.13 & \underline{0.59}\,/\,\underline{1.12} \\
\cmidrule(lr){1-7}
\multirow{3}{*}{Point-grounded} & ShapeR~\cite{siddiqui2026shaper} & \textbf{0.15}\,/\,\textbf{0.51} & 0.48\,/\,1.42 & \textbf{0.39}\,/\,\textbf{1.64} & \underline{1.12}\,/\,\underline{2.57} & 0.86\,/\,5.31 \\
 & HY3D-Omni~\cite{hunyuan3d2025omni} & 0.35\,/\,0.98 & 1.30\,/\,2.28 & -- & -- & -- \\
 & Trac3R (ours) & 0.28\,/\,\underline{0.85} & \underline{0.37}\,/\,\underline{0.95} & \underline{0.92}\,/\,\underline{2.39} & \textbf{1.04}\,/\,\textbf{2.50} & \textbf{0.45}\,/\,\textbf{1.01} \\
\bottomrule
\end{tabular}
\caption{\textbf{The CD split into a shape error (CD-oracle) and a placement error (pose cost)}, as plotted in \cref{fig:tradeoff}. Each cell is median\,/\,mean over the cases of the benchmark; lower is better. \emph{Floor}: the CD-oracle of a perfect shape. The synthetic columns pool Toys4K, GSO and OmniObject3D. --: no run on that benchmark. \textbf{Bold}: best; \underline{underline}: second best.}
\label{tab:tradeoff}
\end{table*}

The CD (Chamfer distance) that ranks the methods in \cref{tab:synthetic,tab:handal,tab:bop,tab:scene} is low only if the shape is right and in the right place. To see which of the two errors each method makes, we split the CD of every case into two parts. \emph{CD-oracle} is the CD after the similarity transform that best aligns the predicted shape to the ground truth shape, found by multi-start ICP, which removes the impact of pose estimation and only evaluates the \emph{shape quality}. \emph{Pose cost} is the CD minus the CD-oracle of the same case; it scores the error caused by sub-optimal placement, and it is zero for a perfectly placed shape. We use pose cost, instead of distance to a ground truth pose label, as the \emph{pose metric}, because it does not suffer from the pose ambiguity of objects with symmetric shapes. Both are in the unit of the CD, and lower is better. For the synthetic objects, we average over Toys4K~\cite{stojanov2021using}, GSO~\cite{downs2022googlescannedobjectshighquality} and OmniObject3D~\cite{wu2023omniobject3d}, separately for GT and MoGe-2~\cite{wang2026moge} depth. \Cref{fig:tradeoff} places every method in the plane of the two errors, using means, which add up to the mean CD (the medians do not). \Cref{tab:tradeoff} gives the medians and means; the text cites medians.

The CD compares two finite samplings of surfaces. Two samplings of the same ground-truth surface therefore give a CD above zero. We call it the \emph{floor} of CD-oracle, which is the lowest CD a perfect shape can reach. 

\paragraph{Overview.}
\Cref{fig:tradeoff} shows the same pattern on all five benchmarks. Trac3R lies in the bottom-left corner: no other method has both a better shape and a better placement. The direct-regression method (green) and the pixel-grounded methods (blue) lie to the right of Trac3R, meaning that their placement is worse, despite some have better shape accuracy in certain tests. The point-grounded baselines (yellow) lie above Trac3R: their shapes are worse, while their placement is close to that of Trac3R in most panels. Trac3R thus has a shape quality close to that of the direct-regression and pixel-grounded methods, and a placement quality close to that of the point-grounded methods.

\begin{figure*}[t]
\centering
\includegraphics[width=\textwidth]{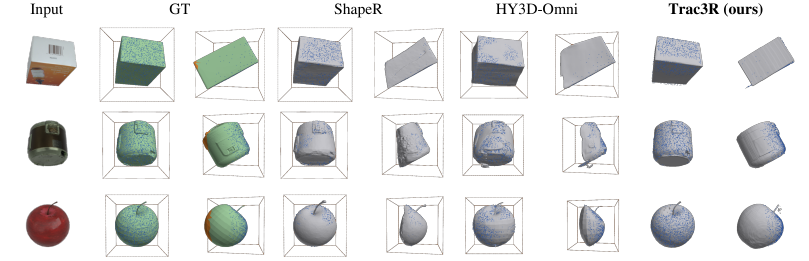}
\caption{\textbf{Why the point-grounded baselines lose shape quality.} ShapeR and Hunyuan3D-Omni scale the observed points (blue) to fill their generation volume (wire cube), so the volume covers only the visible part of the object. For these shown objects, the hidden part of the ground truth shape extends beyond the volume (orange: the GT surface outside the volume of ShapeR). The two baselines squeeze the complete shape into the volume, while Trac3R generates in an object-centric frame and completes the hidden part. Each method is shown from the camera and from a view turned by 80$^\circ$.}
\label{fig:qual_oob}
\end{figure*}

\paragraph{Trac3R against the other point-grounded methods: better shapes.}
Trac3R, ShapeR and Hunyuan3D-Omni all take the observed points as input, but they generate the shape in different frames. ShapeR and Hunyuan3D-Omni~\cite{hunyuan3d2025omni} scale the observed points to fill their generation volume, so the volume covers only the visible part of the object. When the hidden part extends beyond it, the complete shape does not fit, and these methods squeeze it into the volume (\cref{fig:qual_oob}). Trac3R generates the shape in an object-centric frame, so the visible part does not limit the shape. \Cref{tab:oob} splits the synthetic objects by whether the complete ground-truth object fits in the volume of ShapeR. When it fits, ShapeR and Trac3R have shapes of similar quality. When it does not, the shapes of ShapeR and Hunyuan3D-Omni degrade much more than the shape of Trac3R, which degrades as little as those of the pixel-grounded methods. With MoGe-2 depth, more objects do not fit, which is one reason why the point-grounded baselines lose more shape quality than Trac3R with predicted depth. On the placement, Trac3R is on par with ShapeR (\cref{tab:tradeoff}).

\paragraph{Trac3R against the direct-regression and pixel-grounded methods: better placement.}
Trac3R reads its pose from the traced tokens, which are tied to the observed points. These methods estimate the placement in a separate step: SAM~3D~\cite{chen2026sam} regresses it, and the pixel-grounded CUPID~\cite{huang2026cupid}, Pixal3D~\cite{li2026pixal3d} and Mira-Scene fit their output to the depth. On the synthetic objects with GT depth, for example, the median pose cost of Trac3R is 0.28, against 4.82 for SAM~3D. Only Pixal3D places as well as Trac3R, on the isolated, fully visible synthetic objects, where fitting the scale and the distance along the ray suffices; on the real objects and the scenes, Trac3R places better. On the shape, SAM~3D is ahead of Trac3R, most clearly on real images (4.68 against 5.64\,mm on HANDAL~\cite{guo2023handal}, 3.92 against 5.05\,mm on BOP~\cite{hodan2018bop,hodan2024bop}). This shows the room left for improving the shape quality of Trac3R.

\begin{table}[t]
\centering
\footnotesize
\setlength{\tabcolsep}{3pt}
\resizebox{\columnwidth}{!}{%
\begin{tabular}{l l ccc ccc}
\toprule
 & & \multicolumn{3}{c}{GT depth} & \multicolumn{3}{c}{MoGe-2 depth} \\
\cmidrule(lr){3-5}\cmidrule(lr){6-8}
Group & Method & fits & does not fit & $\Delta$ & fits & does not fit & $\Delta$ \\
 & {\scriptsize cases} & {\scriptsize 838} & {\scriptsize 324} &  & {\scriptsize 515} & {\scriptsize 635} \\
\midrule
\multirow{3}{*}{\begin{tabular}{@{}l@{}}Point-\\grounded\end{tabular}} & ShapeR~\cite{siddiqui2026shaper} & \textbf{2.25} & 3.78 & +1.53 & 2.61 & 4.56 & +1.95 \\
 & HY3D-Omni~\cite{hunyuan3d2025omni} & 3.01 & 5.84 & +2.83 & 3.28 & 6.63 & +3.34 \\
 & Trac3R (ours) & \underline{2.32} & \textbf{2.75} & \underline{+0.42} & \underline{2.31} & \textbf{3.01} & \textbf{+0.70} \\
\midrule
\multirow{3}{*}{\begin{tabular}{@{}l@{}}Pixel-\\grounded\end{tabular}} & CUPID~\cite{huang2026cupid} & 2.85 & 4.12 & +1.27 & 2.63 & 3.74 & +1.11 \\
 & Pixal3D~\cite{li2026pixal3d} & 2.57 & 3.13 & +0.56 & 2.42 & 3.15 & +0.73 \\
 & Mira-Scene~\cite{sun2026mira} & 2.56 & \underline{2.95} & \textbf{+0.40} & \textbf{2.29} & \underline{3.02} & \underline{+0.72} \\
\bottomrule
\end{tabular}}
\caption{\textbf{Shape quality on the synthetic objects, split by whether the complete object fits in the generation volume of ShapeR}, a unit cube that the visible points are normalized to (the situation of \cref{fig:qual_oob}). \emph{Fits}: the whole GT surface lies inside this cube; \emph{does not fit}: more than 5\% of it lies outside; $\Delta$: the difference between the former two columns, which is the quality that a method loses when the object does not fit. The point-grounded baselines lose the most, and Trac3R loses as little as the pixel-grounded methods. The split is the same for every method, and each method keeps its own normalization. Each cell is the mean CD-oracle (shape error, \cref{sec:appendix_tradeoff}), in \%; lower is better. \textbf{Bold}: best; \underline{underline}: second best, per column.}
\label{tab:oob}
\end{table}

\paragraph{Trac3R against its base model.}
Trac3R is fine-tuned from the Hunyuan3D-2.1 family~\cite{hunyuan3d2025hunyuan3d21} (\cref{sec:appendix_impl}). Registering the shapes of Hunyuan3D-2.1 to the depth (Hunyuan3D-2.1 + Reg.) gives a shape close to that of Trac3R, but a much worse placement: a mean pose cost of 4.24 against 0.85 on the synthetic objects with GT depth. The gain of Trac3R over its base model thus lies in the placement.

\paragraph{Qualitative results.}
\label{sec:appendix_qual}
\Cref{fig:supp_qual_obj,fig:supp_qual_frames_bop,fig:supp_qual_frames_bs} show more single objects and full frames, in the layouts of \cref{fig:qual_synthetic,fig:qual_frames}. 
The color of the estimated mesh indicates the error between the estimated shape at the estimated pose and the ground truth shape at the ground truth pose. 

\section{Runtime breakdown}
\label{sec:appendix_runtime}

\begin{figure*}[t]
\centering
\includegraphics[width=\textwidth]{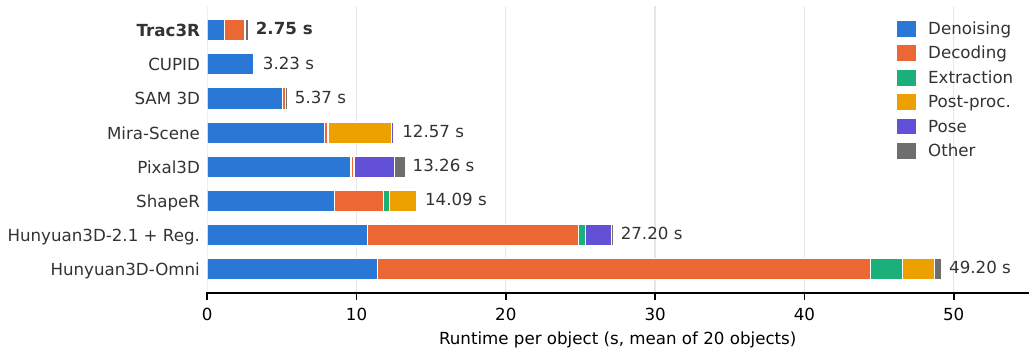}
\caption{\textbf{Where the runtime goes.} Each bar is the mean runtime of one method per object, split into the stages that \cref{sec:appendix_runtime} defines. The denoising segment of every baseline is longer than the complete Trac3R bar. \Cref{tab:runtime} reports the numbers.}
\label{fig:runtime}
\end{figure*}

\begin{table*}[t]
\centering
\footnotesize
\setlength{\tabcolsep}{3pt}
\begin{tabular}{l l c ccccccc}
\toprule
Method & Latent & Denoisers & Denoising & Decoding & Extraction & Post-proc. & Pose & Other & Total \\
\midrule
SAM 3D~\cite{chen2026sam} & Voxels & 2 & 4.69\,/\,5.03 & \underline{0.13}\,/\,\underline{0.20} & 0.031\,/\,0.033 & 0.004\,/\,0.004 & \underline{0.002}\,/\,\underline{0.002} & 0.10\,/\,0.10 & 4.96\,/\,5.37 \\
CUPID~\cite{huang2026cupid} & Voxels & 2 & \underline{2.93}\,/\,\underline{3.10} & \textbf{0.042}\,/\,\textbf{0.050} & \underline{0.025}\,/\,\underline{0.027} & -- & 0.013\,/\,0.015 & \textbf{0.026}\,/\,\textbf{0.034} & \underline{3.05}\,/\,\underline{3.23} \\
Pixal3D~\cite{li2026pixal3d} & Voxels & 3 & 9.23\,/\,9.63 & 0.19\,/\,\underline{0.20} & \textbf{0.004}\,/\,\textbf{0.004} & 0.006\,/\,0.007 & 2.51\,/\,2.68 & 0.71\,/\,0.74 & 12.37\,/\,13.26 \\
Mira-Scene~\cite{sun2026mira} & Voxels & 2 & 7.71\,/\,7.88 & 0.16\,/\,0.21 & 0.034\,/\,0.035 & 4.42\,/\,4.21 & 0.17\,/\,0.13 & 0.10\,/\,0.10 & 12.58\,/\,12.57 \\
\midrule
ShapeR~\cite{siddiqui2026shaper} & VecSet & 1 & 8.51\,/\,8.53 & 3.28\,/\,3.27 & 0.36\,/\,0.38 & 1.79\,/\,1.84 & \textbf{0.001}\,/\,\textbf{0.001} & \underline{0.06}\,/\,\underline{0.06} & 14.07\,/\,14.09 \\
HY3D-Omni~\cite{hunyuan3d2025omni} & VecSet & 1 & 11.43\,/\,11.41 & 33.04\,/\,33.03 & 2.17\,/\,2.15 & 2.19\,/\,2.14 & 0.003\,/\,0.003 & 0.48\,/\,0.47 & 49.36\,/\,49.20 \\
HY3D-2.1~\cite{hunyuan3d2025hunyuan3d21} + Reg. & VecSet & 1 & 10.75\,/\,10.73 & 14.11\,/\,14.11 & 0.51\,/\,0.51 & -- & 1.84\,/\,1.71 & 0.13\,/\,0.13 & 27.35\,/\,27.20 \\
\midrule
Trac3R (ours) & VecSet & 1 & \textbf{1.16}\,/\,\textbf{1.16} & 1.33\,/\,1.32 & 0.06\,/\,0.06 & -- & \textbf{0.001}\,/\,\textbf{0.001} & 0.21\,/\,0.21 & \textbf{2.75}\,/\,\textbf{2.75} \\
\bottomrule
\end{tabular}
\caption{\textbf{Runtime per object, by stage}, in seconds (median\,/\,mean over the same 20 GSO objects; lower is better). \Cref{sec:appendix_runtime} defines the six stages and the timing setup. \emph{Latent}: the kind of 3D latent that the method generates. \emph{Denoisers}: the number of flow or diffusion models that run in sequence for one object. \emph{Total}: the sum of the six stages on each object. --: the method has no such stage. \textbf{Bold:} best; \underline{underline}: second best (Post-proc.\ is not ranked). }
\label{tab:runtime}
\end{table*}

\begin{table*}[t]
\centering
\footnotesize
\setlength{\tabcolsep}{4pt}
\begin{tabular}{l l l cc}
\toprule
Method & Generated latent & Decoded grid & Denoising (s) & Decoding (s) \\
\midrule
\multicolumn{5}{l}{\emph{Voxels}} \\
SAM 3D~\cite{chen2026sam} & $16^3{\times}8$, then $64^3{\times}8$ & $256^3$, occupied voxels & 4.69\,/\,5.03 & 0.13\,/\,0.20 \\
CUPID~\cite{huang2026cupid} & $16^3{\times}16$, then $64^3{\times}8$ & $256^3$, occupied voxels & 2.93\,/\,3.10 & 0.042\,/\,0.050 \\
Pixal3D~\cite{li2026pixal3d} & $16^3{\times}8$, then $32^3{\times}32$, then $64^3{\times}32$ & $1024^3$, occupied voxels & 9.23\,/\,9.63 & 0.19\,/\,0.20 \\
Mira-Scene~\cite{sun2026mira} & $16^3{\times}8$, then $64^3{\times}8$ & $256^3$, occupied voxels & 7.71\,/\,7.88 & 0.16\,/\,0.21 \\
\midrule
\multicolumn{5}{l}{\emph{VecSet}} \\
ShapeR~\cite{siddiqui2026shaper} & $8192{\times}64$ & $257^3$ dense (17.0\,M points) & 8.51\,/\,8.53 & 3.28\,/\,3.27 \\
HY3D-Omni~\cite{hunyuan3d2025omni} & $4096{\times}64$ & $513^3$ dense (135.0\,M points) & 11.43\,/\,11.41 & 33.04\,/\,33.03 \\
HY3D-2.1~\cite{hunyuan3d2025hunyuan3d21} + Reg. & $4096{\times}64$ & $385^3$ dense (57.1\,M points) & 10.75\,/\,10.73 & 14.11\,/\,14.11 \\
Trac3R (ours) & $512{\times}64$ & $193^3$ dense (7.2\,M points) & 1.16\,/\,1.16 & 1.33\,/\,1.32 \\
\bottomrule
\end{tabular}
\caption{\textbf{The latent that each method generates and the grid that it decodes}, which explain the denoising and decoding times of \cref{tab:runtime} (last two columns). Among the VecSet methods, Trac3R generates the smallest latent and decodes the coarsest grid. \emph{Generated latent}: for the voxel methods, the dense first-stage grid (cells per side${}^3 \times$ channels), then the sparse latent grid(s), of which only the occupied cells are generated; for the VecSet methods, tokens $\times$ channels. \emph{Decoded grid}: the grid that the surface is extracted from. \Cref{sec:appendix_runtime} describes the two kinds of latent. Values are read from the released configurations and the timed runs.}
\label{tab:runtime_specs}
\end{table*}

\begin{table}[t]
\centering
\footnotesize
\setlength{\tabcolsep}{2.5pt}
\begin{tabular}{l c ccc}
\toprule
Grid decoding & Queries (M) & Denoising & Dec.\ + extr. & Total \\
\midrule
Dense $193^3$ & 7.19 & 1.16\,/\,1.16 & 1.39\,/\,1.38 & 2.75\,/\,2.75 \\
Dense $385^3$ & 57.07 & 1.19\,/\,1.19 & 11.38\,/\,11.35 & 12.98\,/\,12.97 \\
Hierarchical $385^3$ & 4.01\,/\,4.15 & 1.15\,/\,1.15 & 0.97\,/\,1.03 & 2.43\,/\,2.50 \\
\bottomrule
\end{tabular}
\caption{\textbf{A finer grid does not make a slower Trac3R.} Runtime of Trac3R per object with three ways to decode the grid (seconds, median\,/\,mean, as in \cref{tab:runtime}). Dense $193^3$ is the default. Hierarchical decoding (see \cref{sec:appendix_runtime}) reaches $385^3$ with fewer queries than the default, and its meshes have the same number of faces as the dense $385^3$ meshes on all 20 objects. \emph{Queries}: grid points at which the geometry decoder is evaluated, in millions. \emph{Dec.\ + extr.}: decoding and marching cubes. \emph{Total} also includes the pose and the other stages, so the columns do not add up to it.}
\label{tab:runtime_octree}
\end{table}

\Cref{tab:runtime} and \cref{fig:runtime} split the runtimes of \cref{fig:teaser} into stages. Trac3R takes 2.75\,s per object and is faster than every baseline on all 20 objects, for three reasons:
\begin{itemize}
\item \emph{Denoising is fast.} Denoising is the stage that generates the latent. Trac3R runs one denoiser on 512 tokens in 1.16\,s. For every baseline, denoising alone takes longer than the complete Trac3R pipeline, on all 20 objects.
\item \emph{Pose and post-processing cost almost nothing.} The pose takes 1\,ms, and Trac3R has no post-processing.
\item \emph{Decoding is not a bottleneck.} It takes 1.33\,s at the default resolution. With hierarchical decoding, Trac3R reaches a $385^3$ grid in 2.43\,s in total.
\end{itemize}

\paragraph{Setup.}
We time every method at batch size 1 on the same 20 GSO objects, after one warm-up object, on one NVIDIA RTX PRO 6000 Blackwell Max-Q GPU; stages that run on the CPU use an AMD Threadripper PRO 9975WX. Each method runs the settings of its GSO results in \cref{tab:synthetic}, and the GPU is synchronized before and after each stage. Model loading is not counted. Stages that only produce appearance (the Gaussian decoder of SAM~3D and Mira-Scene, the texture model of Pixal3D) are removed, because the evaluated mesh does not use them. The numbers in the text are medians. The tables also give the means, which lead to the same conclusions.

\paragraph{Stages.}
We divide every pipeline into six stages, which are the columns of \cref{tab:runtime} and the segments of \cref{fig:runtime}. \emph{Denoising} generates the latent: it covers the condition encoders and all denoisers, each a flow or diffusion model that runs through all of its denoising steps. \emph{Decoding} evaluates the latent on a grid: it covers the latent decoders and, on a dense grid, the queries of the geometry decoder at every grid point. \emph{Extraction} builds the mesh: marching cubes on a dense grid, FlexiCubes or a dual grid on the occupied voxels. \emph{Post-processing} is floater removal, hole filling and decimation. \emph{Pose} places the shape in the camera frame. \emph{Other} is data loading, pre-processing and export. The total is the sum of the six stages on each object.

\paragraph{Two kinds of latent.}
The methods generate one of two kinds of 3D latent, which divide the work differently between denoising and decoding. The voxel methods (SAM~3D, CUPID, Pixal3D and Mira-Scene) all use the two-stage latent of TRELLIS~\cite{xiang_structured_2025}. A first denoiser generates a dense, coarse voxel grid that marks the occupied voxels, and a second denoiser generates sparse latents on the occupied voxels only; Pixal3D generates these latents twice, at a coarse and then at a fine resolution. The decoder then evaluates only the occupied voxels. This sparsification keeps the second denoiser and the decoder efficient, at the cost of an additional denoiser. A VecSet latent~\cite{zhang_3dshape2vecset_2023} (ShapeR, Hunyuan3D-2.1, Hunyuan3D-Omni and Trac3R) is a set of tokens of fixed size, so one denoiser suffices; but the geometry decoder is queried at every point of a dense grid, and the decoding takes longer. Resolutions are not directly comparable between the two kinds, because the cost also depends on the size of the latent and of the decoder. Within each kind they are, and \cref{tab:runtime_specs} lists the latent and the decoded grid of every method.

\paragraph{Denoising.}
Trac3R denoises in 1.16\,s. Among the VecSet methods, the denoising time follows the number of tokens: Trac3R denoises 512 tokens in 50 steps, the two Hunyuan3D models denoise 4{,}096 tokens in 50 steps (10.75\,s and 11.43\,s), and ShapeR denoises 8{,}192 tokens in 25 steps (8.51\,s). The voxel methods need 2.93\,s to 9.23\,s for their two or three denoisers. CUPID, SAM~3D and Mira-Scene generate latents of almost the same size (CUPID adds 8 channels for its pixel coordinates in the first stage), so their differences (2.93\,s, 4.69\,s and 7.71\,s) come from their networks and denoising settings. Pixal3D generates the largest latents, with three denoisers, and needs 9.23\,s.

\paragraph{Pose and post-processing.}
Trac3R reads the correspondences from the generated tokens and solves a closed-form fit~\cite{umeyama1991least}, in 1\,ms. SAM~3D regresses the pose and decodes it in 2\,ms. ShapeR and Hunyuan3D-Omni generate the shape in a normalized frame of the points, so their pose stage only inverts this normalization, in 1\,ms and 3\,ms. The other methods also fit the pose after generation, and the cost depends on what the fit uses. Like Trac3R, CUPID and Mira-Scene fit correspondences that the generator outputs, in 13\,ms and 0.17\,s. Pixal3D and Hunyuan3D-2.1 + Reg.\ must first find correspondences on the generated mesh: Pixal3D renders the mesh to fit it to the depth (2.51\,s), and Hunyuan3D-2.1 + Reg.\ matches FPFH features~\cite{rusu2009fast} between the mesh and the observed points with RANSAC before ICP (1.84\,s). Trac3R exports the extracted mesh without clean-up, whereas Mira-Scene, Hunyuan3D-Omni and ShapeR spend 4.42\,s, 2.19\,s and 1.79\,s on post-processing.

\begin{table*}[t]
\centering
\small
\setlength{\tabcolsep}{6pt}
\begin{tabular}{l c c c c}
\toprule
 & \multicolumn{2}{c}{OmniObject3D, CD (\%)} & YCB-V, CD (mm) & HOPE, CD (mm) \\
\cmidrule(lr){2-3}\cmidrule(lr){4-4}\cmidrule(lr){5-5}
Variant & GT depth & MoGe-2 depth & Sensor depth & Sensor depth \\
\midrule
W/o pixel features on tracers & 3.32\,/\,4.05 & 3.35\,/\,3.93 & 5.10\,/\,6.60 & 4.78\,/\,6.47 \\
Trained with ground-truth depth & \textbf{2.25}\,/\,\textbf{2.52} & 3.19\,/\,3.97 & 6.17\,/\,7.45 & 5.54\,/\,6.91 \\
\midrule
Default & 3.02\,/\,3.68 & \textbf{3.16}\,/\,\textbf{3.83} & \textbf{4.68}\,/\,\textbf{5.85} & \textbf{4.62}\,/\,\textbf{5.97} \\
\bottomrule
\end{tabular}
\caption{\textbf{Ablation of two design choices.} Each variant changes one choice of the default setting (last row) and is trained with the same schedule. Without the pixel features on the tracers (\cref{sec:method_tracing}), the error increases on all four test sets. Training with GT depth gives the lowest error when the test depth is also GT, but the highest error with real sensor depth. CD as median\,/\,mean; lower is better. \textbf{Bold}: best per column.}
\label{tab:ablation}
\end{table*}

\begin{figure*}[t]
\centering
\includegraphics[width=\textwidth]{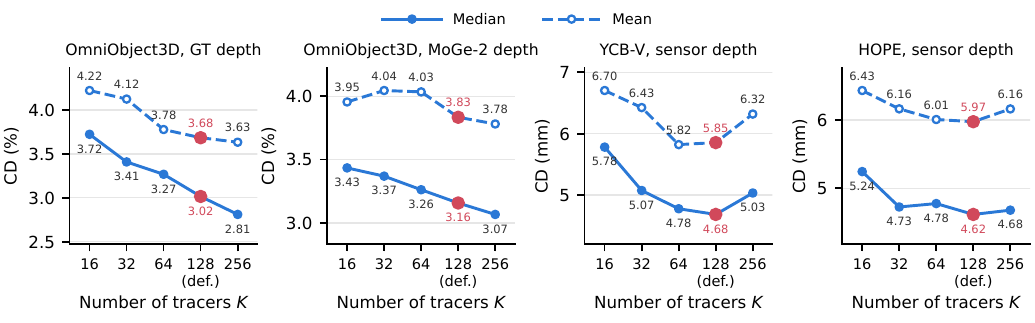}
\caption{\textbf{Effect of the number of tracers $K$} (\cref{sec:method_tracing}), on the test sets of \cref{tab:ablation}. Each point is a model that is trained and tested with that $K$, with other configurations kept the same as the default setting ($K=128$). With fewer tracers than the default, the median CD is higher on all four sets. With 256 tracers, the CD decreases on the synthetic OmniObject3D but increases on the real YCB-V and HOPE.}
\label{fig:tracer_ablation}
\end{figure*}

\paragraph{Decoding.}
Trac3R decodes more slowly than the voxel methods: 1.33\,s against at most 0.19\,s. The voxel methods sparsify their latent between the two denoisers for efficiency, so that the second denoiser and the decoder process only the occupied voxels. A VecSet latent is already compact (512 tokens in Trac3R), so its denoising needs no such step, and the dense grid enters only at decoding, which can also be accelerated by hierarchical decoding (see below). Even with the dense default grid, denoising and decoding together take 2.49\,s in Trac3R, against 2.98\,s in CUPID, the fastest baseline, and 4.81\,s to 44.47\,s in the others. Among the voxel methods, Pixal3D decodes the finest grid ($1024^3$) and takes the longest (0.19\,s). Among the VecSet methods, the decoding time follows the number of grid points: 1.33\,s for 7.2 million points ($193^3$, Trac3R), 3.28\,s for 17.0 million ($257^3$, ShapeR), 14.11\,s for 57.1 million ($385^3$, Hunyuan3D-2.1) and 33.04\,s for 135.0 million ($513^3$, Hunyuan3D-Omni). This is 0.18\,s to 0.25\,s per million points in all four, so the grid resolution, and not the method, sets the decoding cost.

A finer grid does not make Trac3R slow (\cref{tab:runtime_octree}). A dense $385^3$ grid takes 12.98\,s in total. Hierarchical decoding instead evaluates a coarse grid and refines only the cells near the surface, and marching cubes~\cite{lorensen1987marching} then runs on the evaluated cells. It reaches $385^3$ with 4.0 million queries in 2.43\,s in total, which is faster than the dense $193^3$ default on 15 of the 20 objects, and its meshes have the same number of faces as the dense $385^3$ meshes. Hierarchical decoding thus restricts the work to the surface, as the sparsification of the voxel methods does, but it does so during decoding and needs no additional denoiser.

\paragraph{Speed and accuracy.}
Trac3R is fast in part because its configuration is light: among the VecSet methods, it has the most compact latent and the coarsest default grid. All accuracy results of this paper use this configuration, with the dense $193^3$ grid. With it, Trac3R is the most accurate method on the object benchmarks and on par with the scene-specific Mira-Scene on scenes (\cref{tab:synthetic,tab:handal,tab:bop,tab:scene}); \cref{sec:appendix_tradeoff} shows that no other method has both a better shape and a better placement. A compact latent and a coarse grid thus suffice for an accurate shape and an accurate placement, and hierarchical decoding gives a finer mesh when an application needs one.

\section{Ablation study}
\label{sec:exp_ablation}

\Cref{tab:ablation} ablates two major design choices on OmniObject3D~\cite{wu2023omniobject3d} with GT and MoGe-2 depth input, and on the YCB-V~\cite{xiang2017posecnn} and HOPE~\cite{tyree2022hope} datasets with sensor depth input. Without the DINOv2~\cite{oquab2024dinov2} feature feeding into each depth anchor (\cref{sec:method_tracing}), a tracer only carries geometric information, and the error increases on all four sets. Training with ground-truth depth instead of the MoGe-2 point map performs the best when the test depth is also ground truth, while the lead disappears with MoGe-2 depth input, and the model significantly underperforms with sensor depth input. The model never sees imperfect observations in training, and does not transfer to real sensor depth.

\Cref{fig:tracer_ablation} varies the number of tracers $K$ on the same data. Below 128, the median error is higher than with the default on all four sets, and highest with 16 tracers; on YCB-V and HOPE, 64 tracers are close to the default. With 256 tracers, the error decreases on OmniObject3D, most with GT depth, but increases on YCB-V and HOPE. Increasing the tracers above 128 thus only help on synthetic data when the visibility is ideal. We use 128 tracers as we prioritize the real-world performance.

\begin{figure*}[p]
\centering
\includegraphics[width=\textwidth]{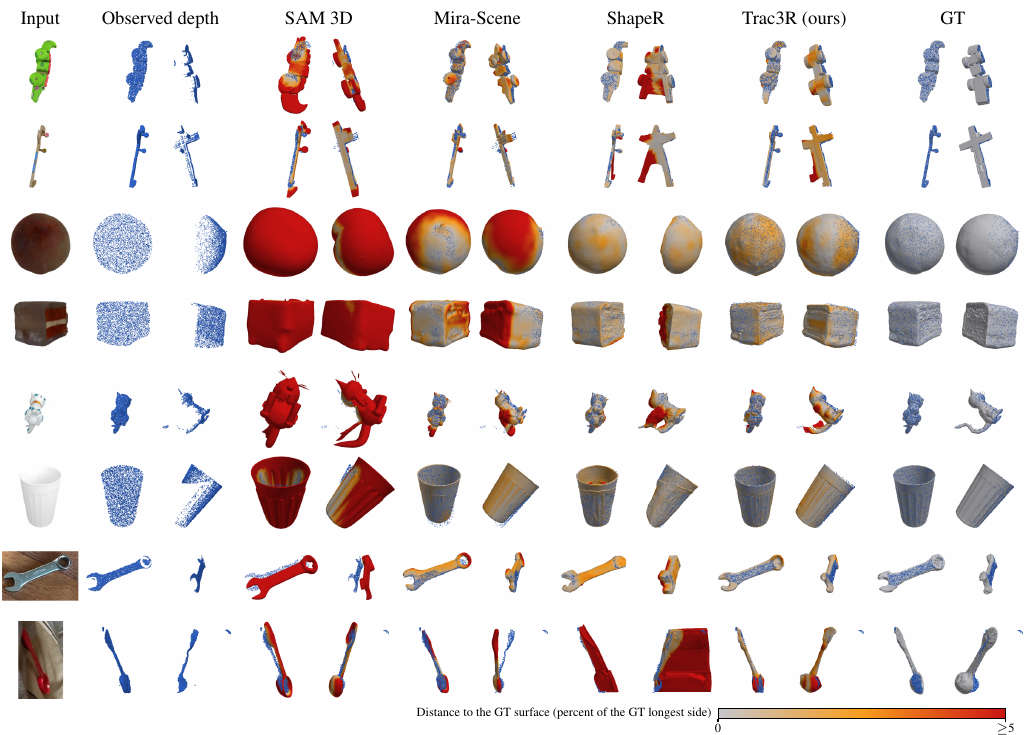}
\caption{\textbf{More single objects.} Rows 1--2: GSO, rows 3--4: OmniObject3D, rows 5--6: Toys4K, all with GT depth. Rows 7--8: real objects with MoGe-2 depth (from HANDAL). Same layout and colors as \cref{fig:qual_synthetic}: every mesh is placed with its predicted pose, with the observed point cloud in blue, and is colored by its distance to the GT surface. Each column shows the camera view and a view turned by 70$^\circ$ around the object.}
\label{fig:supp_qual_obj}
\end{figure*}

\begin{figure*}[p]
\centering
\includegraphics[width=\textwidth]{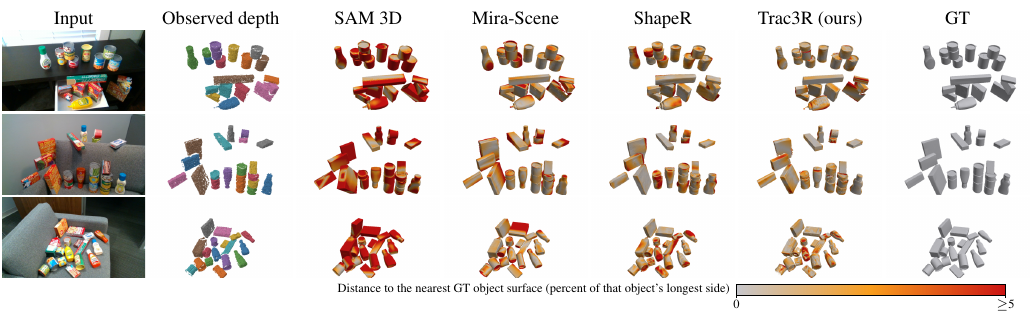}
\caption{\textbf{More real object piles} with sensor depth (from HOPE), visualized from the camera. Same layout and colors as \cref{fig:qual_frames}.}
\label{fig:supp_qual_frames_bop}
\end{figure*}

\begin{figure*}[p]
\centering
\includegraphics[width=\textwidth]{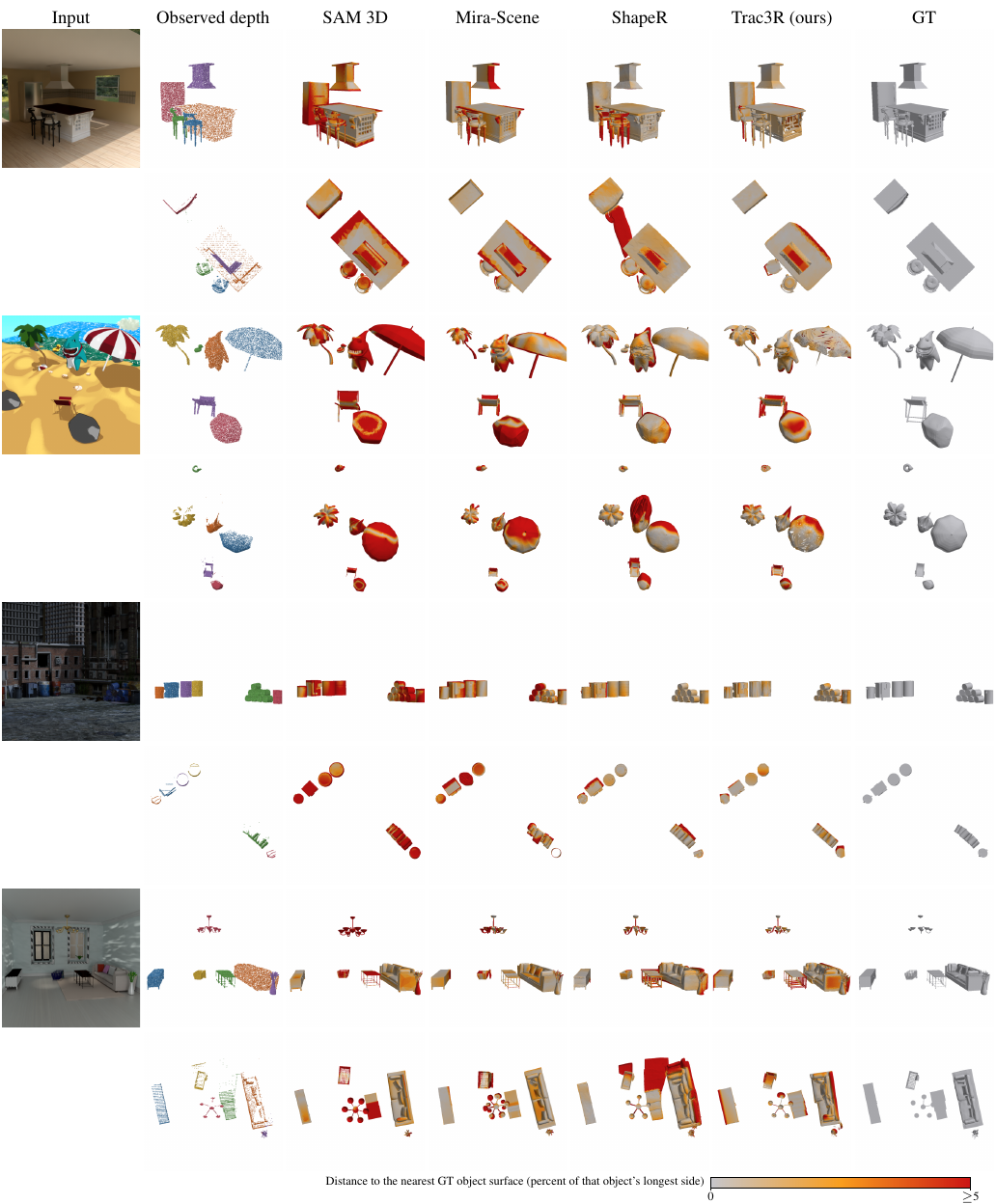}
\caption{\textbf{More scenes} with GT depth (from BlendSwap). Each scene is visualized from the camera (upper row) and from straight above (lower row). Same layout and colors as \cref{fig:supp_qual_frames_bop}.}
\label{fig:supp_qual_frames_bs}
\end{figure*}

\section{Failure case analysis}
\label{sec:appendix_fail}

\begin{figure*}[p]
\centering
\includegraphics[width=\textwidth]{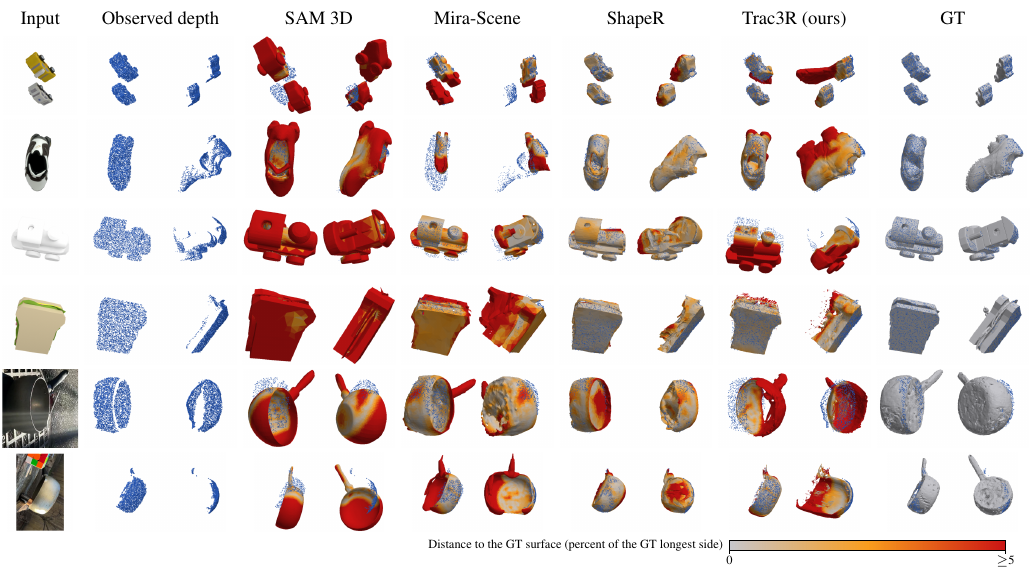}
\caption{\textbf{Failure cases on single objects}: on each row, at least one baseline has a lower CD than Trac3R. \Cref{sec:appendix_fail} discusses the limitations that they show. Rows 1--4: synthetic objects with GT depth (from GSO and Toys4K). Rows 5--6: real objects with MoGe-2 depth (from HANDAL). Same layout and colors as \cref{fig:supp_qual_obj}. }
\label{fig:supp_qual_obj_fail}
\end{figure*}

\begin{figure*}[p]
\centering
\includegraphics[width=\textwidth]{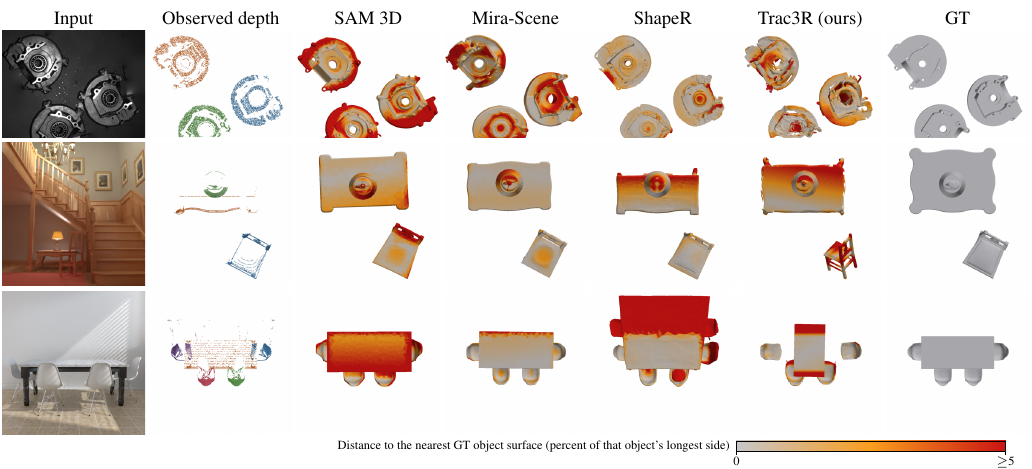}
\caption{\textbf{Failure cases on full frames}: on each row, at least one baseline has a lower mean CD over the objects of the frame than Trac3R (discussed in \cref{sec:appendix_fail}). Row 1: industrial parts with sensor depth (from ITODD), visualized from the camera. Rows 2--3: scenes with GT depth (from BlendSwap), visualized from straight above. Same layout and colors as \cref{fig:supp_qual_frames_bop,fig:supp_qual_frames_bs}.}
\label{fig:supp_qual_frames_fail}
\end{figure*}

\Cref{fig:supp_qual_obj_fail,fig:supp_qual_frames_fail} show failure cases, where at least one baseline has a lower CD than Trac3R. The failures point to four limitations, each of which hints a direction for improvement:
\begin{itemize}
\item \emph{Uncertain hidden side.} The observed points constrain only the visible surface, so the hidden side comes from the object prior of the generator. Where this prior is not strong enough, the generation does not settle on one plausible completion, and the hidden side is incomplete or has artifacts: the boot, the sandwich and the two pots (\cref{fig:supp_qual_obj_fail}, rows 2 and 4--6). A stronger shape prior is the direct remedy. Trac3R generates 512 latent tokens, against 4{,}096 in its base model (\cref{sec:appendix_runtime}), and SAM~3D reconstructs better shapes on real images (\cref{sec:appendix_tradeoff}), which leaves room for a larger latent or a stronger generator.
\item \emph{Missing scene context.} Some hidden extents are ambiguous for the object alone but follow from its surroundings. In \cref{fig:supp_qual_frames_fail} (row 3), the front of the table does not tell how deep the table is, but the wall behind it bounds this depth. Trac3R reconstructs each object from its own crop and points only. Conditioning on the background around the object, or reconstructing the objects of a scene jointly, could resolve such cases.
\item \emph{Wrong placement.} In some cases the shape is right but its placement is wrong: the train (\cref{fig:supp_qual_obj_fail}, row 3) and the chair (\cref{fig:supp_qual_frames_fail}, row 2). The pose is a closed-form fit to the correspondences of the tracers, so a group of tracers that land on the wrong part of the generated shape moves the whole object. A robust fit that rejects inconsistent correspondences, or a check of the placed shape against the observed points followed by a refit, could catch these errors.
\item \emph{Inputs outside the training distribution.} Every training object is one connected shape under one mask. A mask that holds several separate parts never occurs in training, and Trac3R then generates unexpected pieces (\cref{fig:supp_qual_obj_fail}, row 1). Appearance can also be out of distribution: Trac3R underperforms in ITODD, whose grayscale images of metal parts differ from the other datasets, and Trac3R produces fragmented shapes there (\cref{fig:supp_qual_frames_fail}, row 1). Training on multi-part objects and on a wider range of image appearance could address both.
\end{itemize}

\section{Implementation details}
\label{sec:appendix_impl}

\paragraph{Initialization.} The VAE starts from the released Hunyuan3D-2.1 VAE~\cite{hunyuan3d2025hunyuan3d21}. In Stage~1, we add LoRA of rank 8 and scale 16 to its encoder and keep the decoder frozen. The DiT starts from the 2048-token checkpoint of Hunyuan3D-2.1-Small, released with HY3D-Bench~\cite{hunyuan3d2026hy3d}. This DiT generates the latent of the same VAE, and it is smaller than the DiT of the Hunyuan3D-2.1 release.

\paragraph{Token count.} We fine-tune this DiT with 512 shape tokens instead of 2048, for three reasons. First, HY3D-Bench trains the DiT progressively with 512, then 2048 tokens, so 512 tokens are part of the training history of this checkpoint. Second, the VAE has no fixed token count: its encoder takes the number of query points as an input, and our Stage~1 VAE reconstructs shapes from 512 tokens nearly as well as from 2048. Third, the cost of the DiT increases with the number of tokens, so 512 tokens keep training and denoising fast (\cref{sec:appendix_runtime}).

\paragraph{DiT training.} We use AdamW with a batch of 80 per GPU on two GPUs. The learning rate is $5\times10^{-5}$ for the pretrained weights and $5\times10^{-4}$ for the new layers. At step 40k, we decrease both learning rates by a factor of 10, and we use the checkpoint at step 60k. At test time, we sample with 50 Euler steps and a classifier-free guidance scale of 5.

\paragraph{Depth screening.} We screen sensor depth and MoGe-2 point maps, in training and at test time; GT depth is used as it is. (1)~To correct the scale of MoGe-2, which does not recover metric scale reliably, we fit its depth to the reference depth in the mask region by scaling and shifting. (2)~To remove the isolated points that appear at depth edges, statistical outlier removal drops a point when its mean distance to its 16 nearest neighbors is more than the median plus two robust standard deviations. (3)~To remove detached groups of points, such as background seen through the edge of the mask, a cluster screen links points closer than six times the median point spacing and keeps every component whose nearest point lies within half the extent of the largest component. (4)~To handle objects with little valid sensor depth, where the largest component can be the background inside the mask, a fallback runs, on sensor depth only, if the cluster screen would keep less than 70\% of the points. Because the background is always behind the object, the fallback sorts the points by depth instead, splits them where the depth increases by more than 5\% of the median depth, and keeps the nearest group. To skip stray points in front of the object, this group must hold at least 5\% of the points. MoGe-2 point maps are dense and do not need the fallback. A guard keeps all points when fewer than 64 would remain.

\fi

\end{document}